\documentclass[lettersize,journal]{IEEEtran}
\usepackage{amsmath,amsfonts}
\usepackage{algorithmic}
\usepackage{algorithm}
\usepackage{array}
\usepackage[caption=false,font=normalsize,labelfont=sf,textfont=sf]{subfig}
\usepackage{textcomp}
\usepackage{stfloats}
\usepackage{url}
\usepackage{verbatim}
\usepackage{graphicx}
\usepackage{cite}
\usepackage{amssymb}

\usepackage{booktabs}
\usepackage{multirow}
\usepackage{float}
\usepackage{enumitem}

\usepackage{soul}
\usepackage{color}
\usepackage{cite}

\usepackage{subfig}

\usepackage{tabularx}

\usepackage{arydshln}

\usepackage{orcidlink}

\newcolumntype{C}{>{\centering\arraybackslash}X} 

\newcommand{\eg}{\textit{e.g.}}
\newcommand{\ie}{\textit{i.e.}}
\newcommand{\etal}{\textit{et al.}}
\newcommand{\cf}{\textit{cf. }}

\begin{document}

\title{Domain Expansion and Boundary Growth for Open-Set Single-Source Domain Generalization}

\author{Pengkun~Jiao \orcidlink{0009-0007-0542-3482},
        Na~Zhao \orcidlink{0000-0003-2329-7014},
        Jingjing~Chen \orcidlink{0000-0003-3148-264X}
        and Yu-Gang~Jiang \orcidlink{0000-0002-1907-8567}, \textit{Fellow, IEEE}

\thanks{
This research is supported by the National Natural Science Foundation of China (NSFC) (Grant No. 62072116); the Agency for Science, Technology, and Research (A*STAR) under its MTC Programmatic Funds (Grant No. M23L7b0021); and the Ministry of Education, Singapore, through its MOE Academic Research Fund Tier 1 - SMU-SUTD Internal Research Grant (SMU-SUTD 2023\_02\_09).
\textit{(Corresponding authors: Jingjing Chen; Na Zhao.)}
}

\thanks{Pengkun Jiao, Jingjing Chen, and Yu-Gang Jiang are with Fudan Vision and Learning Laboratory (FVL), Fudan University, Shanghai, China (e-mail: pkjiao23@fudan.edu.cn; chenjingjing@fudan.edu.cn; ygj@fudan.edu.cn).
Na Zhao is with the Department of Information Systems Technology and Design, Singapore University of Technology and Design (e-mail: na\_zhao@sutd.edu.sg).}}
        



\maketitle

\begin{abstract}
Open-set single-source domain generalization aims to use a single-source domain to learn a robust model that can be generalized to unknown target domains with both domain shifts and label shifts. The scarcity of the source domain and the unknown data distribution of the target domain pose a great challenge for domain-invariant feature learning and unknown class recognition. In this paper, we propose a novel learning approach based on domain expansion and boundary growth to expand the scarce source samples and enlarge the boundaries across the known classes that indirectly broaden the boundary between the known and unknown classes. Specifically, we achieve domain expansion by employing both background suppression and style augmentation on the source data to synthesize new samples. Then we force the model to distill consistent knowledge from the synthesized samples so that the model can learn domain-invariant information. Furthermore, we realize boundary growth across classes by using edge maps as an additional modality of samples when training multi-binary classifiers. In this way, it enlarges the boundary between the inliers and outliers, and consequently improves the unknown class recognition during open-set generalization. Extensive experiments show that our approach can achieve significant improvements and reach state-of-the-art performance on several cross-domain image classification datasets.
\end{abstract}


\begin{IEEEkeywords}
Open-set, single-source domain generalization, out of distribution, domain expansion, class boundary, edge map, binary classifier.
\end{IEEEkeywords}

\begin{figure}[t] 
\centering
\includegraphics[clip, trim=0cm 0cm 0cm 1.5cm, width=0.47\textwidth]{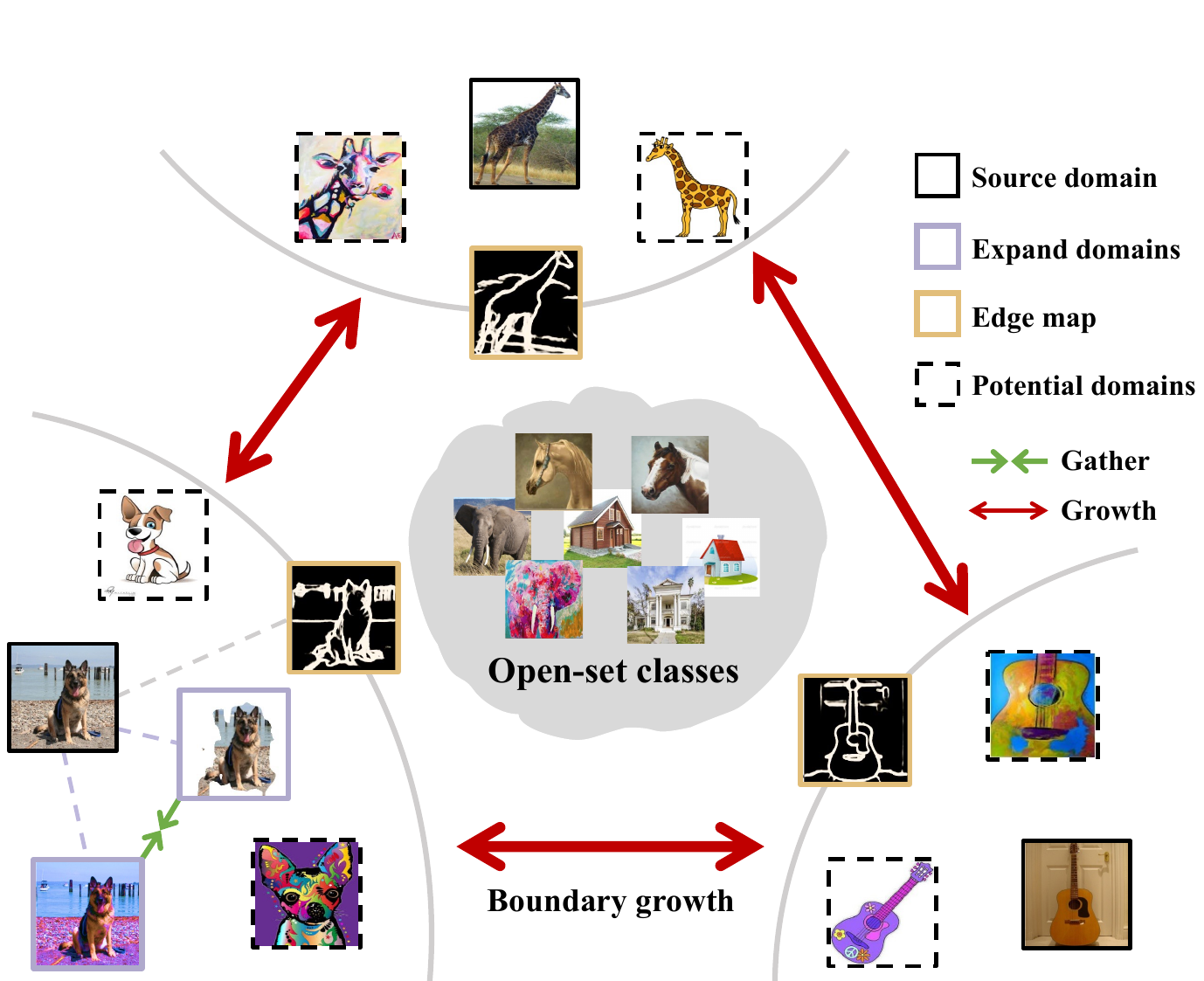} 
\caption{In the open-set single-source domain generalization (OS-SDG) setting, the capacity of our model in recognizing out-of-distribution known and open-set classes can be simultaneously enhanced when the known class distributions in the source feature space are
adequately discriminative and well separated through domain expansion and boundary growth.
}
\label{fig:teaser} 
\end{figure}

\section{Introduction}

\IEEEPARstart{W}{hen} the training and test data are independently and identically distributed (\textit{i.i.d.}), deep neural networks have shown remarkable performances in various computer vision tasks \cite{liu2022convnet_BW, pham2021meta_BW, yu2022coca_BW,jiao2024INHA}. However, the performances will suffer from a catastrophic drop once the training and test data have different distributions \cite{ben2010iid2, vapnik1991iid1,zhao2022synthetic,yang2023auc}. Moreover, these distribution shifts are generally inevitable in real-world scenarios since the collection of training datasets is inherently a sub-sampling of the real-world data; thus, it is impossible to have the same distribution with unseen testing datasets that are usually collected separately under different conditions (\eg~time and space). 
Particularly, distribution shifts can appear as both domain shifts \cite{stacke2020domain_shift,zhao2024styleh-vdg} and label shifts \cite{lipton2018labelshift,wang2022openauc,zhao2024robustvr-own} in the unseen testing datasets, resulting in target domains with different domain characteristics and larger label spaces compared to the source domain. 
To generalize to unseen target domains with unknown categories, open-set domain generalization (OS-DG)~\cite{shu2021OpenDomainGeneralization_OS-DG} is studied, with the aim of learning a generalizable model from multiple source domains.

However, gathering annotated data from multiple domains can be both difficult and costly. As a result, a more practical and appealing approach involves using a single-source domain. This has spurred interest in the study of open-set single-source domain generalization (OS-SDG)~\cite{zhu2022crossmatch_SDG},  where a model is trained solely on a single source domain to identify known categories and discern unknown categories in unseen domains. Despite its appeal, this reliance on a singular source domain poses a significantly greater challenge compared to OS-DG.

CrossMatch \cite{zhu2022crossmatch_SDG} is the first work solving OS-SDG. Built upon the adversarial data augmentation method \cite{volpi2018ada_SDG, zhao2020meada_SDG}, CrossMatch generates two kinds of auxiliary training data: samples outside the source data distribution and samples outside the source label space, which are used to extend the source domain and act as unknown classes, respectively. 
However, the quality of samples generated through adversarial training can be unstable, resulting in unrealistic and subpar samples, particularly for unknown classes.
On the other hand, SODG-Net \cite{bele2024SODG} generates pseudo-open samples by combining the immediate features of two known classes. Nevertheless, this approach could potentially undermine the recognition of the known classes, as pseudo-open samples incorporate partial features from these known classes.
Both of these prior works rely on the generation of pseudo-open samples, which may not accurately or sufficiently represent the actual open samples in the unseen domains.

In this paper, we propose a novel \textbf{D}omain \textbf{E}xpansion and \textbf{B}o\textbf{U}ndary \textbf{G}rowth based method, named DEBUG, that only augments source samples in terms of known classes, without generating any pseudo-open samples. 
An intuition stems from the observation that the capacity of the model in recognizing out-of-distribution known and unknown classes can be simultaneously enhanced once the known class distributions in the source feature space are adequately discriminative and separated by a large margin, as illustrated in Figure \ref{fig:teaser}.

The profound rationale for DEBUG is rooted in the theory that inductive bias aids in generalization \cite{xuhong2018explicit_inductive_bias, goyal2022inductive}. Enhancing the induction in semantic attribution is beneficial for generalization to unseen domains \cite{wang2019learning_semantic_tmm, zhang2023atzsl_tmm}. However, training solely on a single source domain may inadvertently introduce a bias tied to domain-specific characteristics, which could adversely affect cross-domain generalization. 
We classify these domain characteristics into two distinct types: class-irrelevant content (primary background information) \cite{pan2019foreground_tmm} and style\cite{huang2017adain}.
To address these biases, we propose two techniques: background suppression (BS) and global probabilistic-based style augmentation (GPSA).
More specifically, we use an off-the-shelf foreground-background segmentation method\footnote{Note that our method is relatively robust to the choice of the unsupervised segmentation methods, as demonstrated in Table~\ref{table:ablation_BS}.} to get a coarse yet free foreground mask for each sample.
Based on the resultant coarse mask, we remove the irrelevant background regions from the original samples, referred to as \textit{background suppression}.  
Additionally, we introduce a global probabilistic-based style augmentation method that globally models the feature statistics (\ie~mean and standard deviation) of the original and background-suppressed samples as Gaussian distributions. We augment both the original and background-suppressed samples by replacing their instance-level style statistics with randomly sampled style statistics from the global probabilistic distributions.
Subsequently, with the paired augmented samples with and without backgrounds, we force the model to learn consistent representations via knowledge distillation. The consistency regularization between two samples with different content and style augmentations promotes domain-invariant representation learning. Moreover, the consistency with the background-suppressed augmentation can reduce the intra-class variance since the removal of a background can mitigate the spurious effect of irrelevant context.

Despite the elimination of inductive biases related to domain characteristics, some class-unspecific features, such as common object parts (\eg, furry tails), may still increase the risk of misclassification, particularly for unknown classes that have not yet had their class-specific features captured.
In response to this, we develop the Boundary Growth strategy, which diminishes the emphasis on common but class-unspecific features, which subsequently leads to further sufficient separation of class distributions and creates more room for open-set recognition.
Specifically, we adopt the multi-binary classifiers \cite{saito2021ovanet} that train a one-vs-all classifier for each class to deal with open-set recognition, and we present a novel strategy based on edge maps that are considered as a new modality of the samples to train the multi-binary classifiers.
Specifically, we extract edge maps from the original samples, treating them as additional samples with the same semantic yet different spreads in the feature space (see Figure 3). When training a binary classifier for a specific class, a boundary is learned between the edge maps of the positive samples and the hard negative samples, either in their original form or as edge maps.

We conduct extensive experiments on a variety of cross-domain datasets, including PACS \cite{li2017pacs}, Office31 \cite{saenko2010office31}, OfficeHome \cite{saenko2010office31}, and DomainNet126 \cite{saito2019domainnet126}. Our proposed DEBUG shows significant and consistent improvements over CrossMatch, especially on unknown class recognition. This demonstrates the effectiveness of our proposed approach for open-set single-source domain generalization.

Our contributions can be summarized as follows:
\begin{itemize}\setlength\itemsep{0.25em}
    \item We offer a novel insight into OS-SDG: the recognition ability of out-of-distribution known and open-set unknown classes can be improved when there is clear discrimination and sufficient separation in the source feature space's distribution of known classes. We achieve the discrimination and separation 
    through our domain expansion and boundary growth design.
    \item We introduce a comprehensive method for expanding the single-source domain: we augment the original samples by suppressing the background and adding style disturbances sampled from global probabilistic style statistics, and guide models to learn content-invariant information by distilling knowledge from these background-suppressed augmentations.
    \item We propose a novel boundary growth technique to preserve spaces for unknown classes encountered in the target domains. We utilize edge maps as an additional modality and integrate them into the training of multi-binary classifiers. The edge maps expand the initial distribution of each class, and the optimization of the binary classifier for each class extends the class boundary of the initial distribution.  
    \item We demonstrate that our proposed approach consistently outperforms existing methods on various cross-domain image datasets in the challenging open-set single-domain generalization setting. 
\end{itemize}

\section{Related Work}

\subsection{Domain Generalization} \label{sec:2.1}
\textit{Domain Generalization} (DG) aims to learn a robust model
that can generalize to unseen target domains. Great progress has been made in DG over the years, and the existing methods can be generally classified into three categories according to the techniques they use: 
1) data augmentation \cite{zhou2020generateNovelDomains_DG_aug,wang2020domain_mixup, zhou2021mixstyle_DG_aug, li2022uncertainty,zhao2022style}, 2) representation learning \cite{ghifary2015multi-task_autoencoders_DG_invariant, zhang2022towards_disentangle_DG_invariant, lv2022causality, muandet2013invariantFeatureRepresentation_DG_invariant, ilse2020diva_DG_invariant, liu2022decompose_tmm, li2022deep_margin_sensitive_tmm}, and 3) learning strategies, \eg, meta-learning \cite{balaji2018metareg_DG_meta, du2020variationalInformationBottleneck_DG_meta, li2018meta_learning_for_DG_meta, li2019eature-critic_networks_DG_meta}, ensemble learning \cite{mancini2018source_specific_nets_DG_ensemble, seo2020domainSpecificNormalization4domainGeneralization_DG_ensemble, wang2020Domain-oriented_feature_embedding_DG_ensemble, xu2023federated_domain_hall_tmm}, self-supervised learning \cite{carlucci2019dg_self_single, bucci2021dg_self_multi, kim2021selfreg} and gradient operation \cite{huang2020dg_grad_self, shi2021dg_grad_gra_match, tian2022dg_grad_neuron}.
In the data augmentation-based methods, Zhou \etal \cite{zhou2020generateNovelDomains_DG_aug} augment the source domains by employing a data generator to synthesize data from pseudo-novel domains. Wang \etal \cite{wang2020domain_mixup} synthesize new samples by mixing up data from multiple source domains. DSU \cite{li2022uncertainty} models the feature statistics within each training batch as a Gaussian distribution to capture domain shift uncertainty, and then samples new features from this estimated distribution to simulate diverse potential domains. 
To learn domain invariant representation, Mu \etal \cite{muandet2013invariantFeatureRepresentation_DG_invariant} learn domain invariant feature representations by a kernel-based method.  
Ilse \etal \cite{ilse2020diva_DG_invariant} propose a domain invariant variational auto-encoder that learns independent latent subspaces for domain, class, and residual variations.
In terms of learning strategies, Du \etal \cite{du2020variationalInformationBottleneck_DG_meta} adopt a probabilistic meta-learning model for domain generalization, where they model shared classifier parameters as distributions to estimate prediction uncertainty and obtain domain-invariant representations using their proposed meta variational information bottleneck principle.
On the other hand, Bucci \etal \cite{bucci2021dg_self_multi} enhance cross-domain robustness by adding self-supervised learning tasks, such as solving jigsaw puzzles and recognizing the image orientation. 
Tian \etal \cite{tian2022dg_grad_neuron} generalize the model to out-of-distribution samples by maximizing the neuron coverage of the model with the gradient similarity regularization between the original data and the augmented data.

\textit{Single Domain Generalization} (SDG) is a special and extreme case of DG, which assumes only one source domain is available. The single-source assumption raises additional challenges compared to multi-source DG. Existing works mainly tackle the SDG problem by synthesizing auxiliary training data to increase the diversity of the training data.
Adversarial data augmentation (ADA) \cite{volpi2018ada_SDG} is a commonly used technique, which generates training samples by backpropagating the inverse gradient of the classifier. 
Maximum-entropy adversarial data augmentation \cite{zhao2020meada_SDG} further improves ADA by generating hard negative samples. Furthermore, ASR-Norm \cite{fan2021adaptiveNormalization_SDG} designs an adaptive normalization scheme based on ADA to improve the model generalization ability across domains. Peng \etal \cite{peng2022out} employs a meta-learning-based scheme to efficiently organize the training of adversarially augmented ``fictitious'' domains.
Unlike these ADA-based methods, L2D \cite{wang2021learning2diversify_SDG} proposes a style-complement module to enhance the generalization power of the model, and a progressive domain expansion network \cite{li2021ProgressiveDomainExpansion} designs a learnable generator to expand the source domain progressively. 
Recently, Cugu \etal \cite{cugu2022attention} randomly applied visual corruptions on the training data, and enforced visual attention consistency between the model's activation maps for the original and corrupted input images.
SimDE \cite{xu2023simde} expands the source domain by generating uncertain samples through a trade-off between entropy maximization and cross-entropy minimization, and enhances the diversity of domain expansions by training a pair of generators that alternate the guidance of dual classifiers.

\subsection{Open-Set Recognition}
\textit{Open-Set Recognition} (OSR) aims to identify ``unknown'' classes that are unseen during training in the testing stage. 
Prior works tackle OSR mainly by: 1) modifying SoftMax for potential unseen classes \cite{bendale2016OSdeep_networks}, 2) generating open-set samples \cite{neal2018osr_conterfactual, kong2021opengan}, 3) employing placeholders for novel class distribution anticipation \cite{zhou2021learning_placeholder}, 4) prototype-based methods \cite{chen2021osr_proto_adversarial}, and 5) enhancing closed-set accuracy \cite{saito2021ovanet, vaze2021AGoodClassifier, jang2022collective_one-vs-rest_OVA}.
In the direction of open-set samples generation, Neal et al. \cite{neal2018osr_conterfactual} use generative adversarial networks to generate examples resembling training data but not belonging to any training category, thus reformulating open-set recognition as a classification task with an additional ``unknown'' class.
OpenGAN \cite{kong2021opengan} creates synthetic open data adversarially at the feature level rather than at the pixel level.
In contrast to these generation-based methods, some methods circumvent the challenging open-set sample generation procedure by improving open-set recognition by enhancing the close-set classification. Vaze \etal \cite{vaze2021AGoodClassifier} demonstrate a strong correlation between closed-set and open-set performance in OSR. They find that efficient identification of closed-set classes contributes to open-set identification.
OVA \cite{jang2022collective_one-vs-rest_OVA} trains a one-vs-all classifier for each class in the closed-set, aiming to reduce intra-class distance and increase inter-class
distance, consequently creating more room for effective rejection of unknown classes.
Unfortunately, these existing OSR methods will fail to identify open-set classes when the test distribution is different from the training one.

\subsection{Open-Set Domain Generalization}
\textit{Open-Set Domain Generalization} (OS-DG) aims to identify unknown classes in target domains in addition to recognizing out-of-distribution known classes as in SDG. 
\cite{shu2021OpenDomainGeneralization_OS-DG} is the first work that introduces the task of OS-DG. It solves the task by first augmenting domains on both the feature-level and label-level via Dirichlet mixup and soft-labeling, and then conducting meta-learning over the domains to improve the generalization ability of the model.
In addition to meta-learning, some other learning strategies have been explored. For example, Katsumata \etal \cite{katsumata2021openDGviaMetricLearning_OS-DG} use metric learning to diffuse feature representations of unknown samples. Shao \etal \cite{shao2022Open-set_learning_under_covariate_shift_OS-DG} propose a supervised contrastive learning based framework to cope with open-set learning under domain shifts.
Yang \etal \cite{yang2022one_ring_OS-DG} propose a new training scheme to learn a ($n$+1)-way classifier to predict $n$ sources and one unknown class, and uses weighted entropy minimization to improve the generalization ability of the model.
MEDIC \cite{wang2023medic} leverages the gradient matching property inherent in meta-learning to establish a well-balanced decision boundary among classes.
Despite the fact that these methods can learn a relatively robust model that generalizes to cross-domain data and identifies unknown classes, their success heavily depends on the richness of multiple source domains, which might not be guaranteed in reality due to the heavy collection and annotation cost of cross-domain datasets.

Similar to the relation between DG and SDG, \textit{Open-set Single-source Domain Generalization} (OS-SDG) is an extreme case of OS-DG. Research on OS-SDG is largely unexplored. The first work is  CrossMatch \cite{zhu2022crossmatch_SDG}, which generates auxiliary training data for both in-distribution classes and open-set classes, based on an adversarial data augmentation technique, to train a generalizable model.
Contrastingly, SODG-NET \cite{bele2024SODG} bolsters generalization by diversifying the styles of known class samples. It generates pseudo-open samples to train a unified multiclass classifier, which is adept at managing both open and closed-set data.
However, both these methods necessitate the creation of pseudo-open samples, which may significantly deviate from realistic open samples.
In this paper, we propose a novel method based on domain expansion and boundary growth, achieving improved and stable performance without the need to generate pseudo-open class samples.

\section{Method} \label{section:method}

In this paper, we study OS-SDG in the context of the image classification task. Given only one source domain $\mathcal{S} =\{(x_{i}^{s},y_{i}^{s}) \}_{i=1}^{N_s}$ containing $N_s$ samples with the label space $ C^s$, our objective is to learn a robust model $\mathcal{F}(.)$ that can generalize to any target domain $\mathcal{T}=\{(x_{i}^{t},y_{i}^{t}) \}_{i=1}^{N_t}$, whose label space $C^t$ is a superset of $C^s$ (\ie~$C^s \subset C^t$). 
Specifically, $\mathcal{F}$ should be able to recognize the target samples from known categories $C^s$, and identify the samples from open-set categories $C^{t\setminus s}$ as unknown, which is formally formulated as:

\begin{eqnarray} \label{eq1}
    \begin{cases}
     \mathcal{F} (x^{t}_i ) = y^{t}_i, & \text{ if  }  y^{t}_i \in C^{s},   \\
     \mathcal{F} (x^{t}_i ) = \text{unknown}, & \text{ if  }  y^{t}_i \notin C^{s}.
    \end{cases}
\end{eqnarray}

\begin{figure*}[htbp] 
\centering
\includegraphics[width=0.9\textwidth]{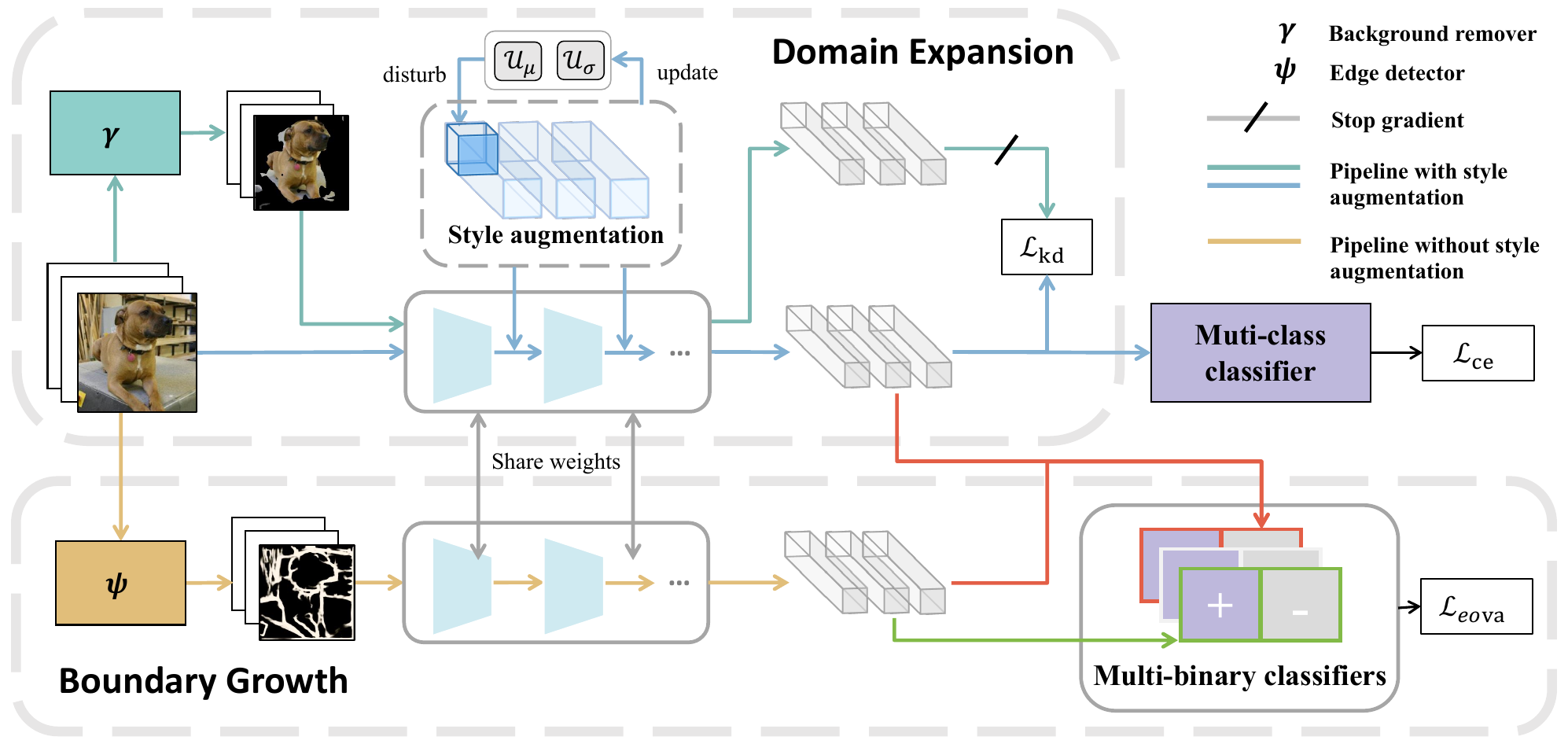} 
\caption{\textbf{The overall framework of our proposed DEBUG}. Domain expansion is employed by applying background suppression and style augmentation and then computing the knowledge distillation loss $\mathcal{L}_{kd}$. Boundary growth is employed by extracting edge maps of the images and then using them as hard positive and negative samples to compute the edge-vs-all loss $\mathcal{L}_{eva}$.
}
\label{fig:framework} 
\end{figure*}

\subsection{Framework Overview}
The framework of our proposed DEBUG is illustrated in Figure \ref{fig:framework}. Our DEBUG consists of three learnable modules: a feature encoder, a linear multi-class classifier, and $|C^s|$ one-vs-all classifiers (\ie~multi-binary classifiers). 
Our DEBUG is trained with the conventional cross-entropy loss $\mathcal{L}_{ce}$ and other two losses (\cf~Eq. \ref{eq:loss_kd} and \ref{eq:loss_eova} for the details) that are computed for the sake of domain expansion and boundary growth, respectively.


\subsection{Domain Expansion}
We expand the limited source domain dataset by augmenting the samples from both the content and style perspectives, using background suppression (Sec.~\ref{sec:bg_suppression}) and global probabilistic-based style augmentation (Sec.~\ref{sec:gp_style_aug}), respectively. With these augmented samples, our model is able to learn domain-invariant features by distilling consistent knowledge that resists both content and style perturbations. 

\subsubsection{Background Suppression} \label{sec:bg_suppression}
The background region in an image is usually irrelevant to the semantic category of the image, but it plays an adverse role in the representation learning \cite{scholkopf2021causalrepresentationlearning}. If a model is not robust enough to disentangle such irrelevant information from the semantic-relevant information, the resultant representation could be susceptible to domain perturbations such as background changes. In other words, disentangling background effects from representation learning can facilitate the acquisition of domain-invariant features.

Motivated by this, we introduce a simple background suppression scheme to coarsely mask out background regions. 
Unlike a few recent works that learn precise binary masks to separate foreground and background via complex adversarial training, we only use an off-the-shelf unsupervised foreground-background segmentation method to obtain coarse binary masks. Specifically, we opt for DenseCLIP \cite{zhou2021denseclip}, an open-vocabulary segmentation approach, due to its simplicity and efficacy.
DenseCLIP is based on the pre-trained multi-modal CLIP \cite{radford2021clip}, and it leverages the powerful semantic encoding capability of CLIP to obtain coarse dense labels (\ie~segmentation result) for an image without training or fine-tuning. 
By adopting this off-the-shelf segmentation tool, we generate a free binary mask for each image according to the semantic class of the image. Subsequently, we remove the background region from the image based on the coarse mask, resulting in a background-suppressed image.



\subsubsection{Global Probabilistic-based Style Augmentation}\label{sec:gp_style_aug}
As pointed out in AdaIN \cite{huang2017adain}, the feature statistics, \ie~channel-wise mean and standard derivation encode the style information, which is one key type of domain shift. Previous works \cite{tang2021selfnorm, zhou2021domain} apply style augmentation by computing all the feature statistics from the whole dataset and then swapping (or mixing) them with the original style feature statistics. Although these methods can generate additional samples, they rely on deterministic feature statistics that do not account for the underlying distribution of the feature statistics.
A recent work DSU \cite{li2022uncertainty} proposes to explore the uncertain statistics by modeling the feature statistics computed within a mini-batch as Gaussian distributions. During batch training, DSU replaces the original statistics of the style features with the new ones sampled from the locally estimated Gaussian distributions. 

Unfortunately, the estimation of batch-based statistics in DSU can be prone to instability and bias, due to fluctuations in the data distribution across different training batches. 
Moreover, the local uncertainty estimation in DSU is insufficient in depicting the global distribution scope of probabilistic feature statistics, which provide more diverse and smooth augmentations. The global probabilistic distribution is important for open-set SDGs since we wish to expand the feature space of the known classes from the source domain (via more diverse augmentations) but meantime leave some spaces in between the known classes for the open-set classes in potential target domains (via smooth augmentations).  
To this end, we propose a global probabilistic modeling method to explore the style uncertainty within the source domain for style augmentation. Instead of adopting a naive way that computes the feature statistics within the whole dataset offline for the probabilistic modeling, we estimate the global probabilistic model in an \textit{online} fashion.

More specifically, for the intermediate feature of a mini-batch $z \in \mathbb{R} ^{B \times C \times H \times W }$ produced by middle layers of the feature encoder, where $B$ is the batch size and $C$ is the number of channel, 
we first compute its instance level channel-wise mean 
$\mu\in \mathbb{R} ^{B\times C}$ and variance $\sigma^2 \in \mathbb{R} ^{B\times C}$:
\begin{gather}
    \mu (z) = \frac{1}{HW} \sum_{h = 1}^{H} \sum_{w = 1}^{W} z|_{h,w}, \\
    \sigma ^{2}  (z) = \frac{1}{HW} \sum_{h = 1}^{H} \sum_{w = 1}^{W} \left ( z|_{h,w} - \mu(z) \right ) ^{2}.
\end{gather}
Then, following the non-parametric uncertainty estimation mechanism as in DSU, we respectively calculate the variance of statistics $\mu(z)$ and $\sigma(z)$ in one batch:
\begin{gather}
    \Sigma_{\mu}^{2}  = \frac{1}{B} \sum_{b = 1}^{B} \left ( \mu(z)|_b - \mathbb{E}_b\left [ \mu(z) \right ]   \right ) ^2, \\
    \Sigma_{\sigma}^{2}  = \frac{1}{B} \sum_{b = 1}^{B} \left ( \sigma(z)|_b - \mathbb{E}_b\left [ \sigma(z) \right ]   \right ) ^2.
\end{gather}
Once obtaining $\Sigma_{\mu}^{2}$ and $\Sigma_{\sigma}^{2}$ of $z$, we use a moving average algorithm to update global uncertainty statistics:
\begin{gather} 
    \mathcal{U}_{\mu}  = \alpha  \mathcal{U}_{\mu} + (1-\alpha ) \Sigma_{\mu}^{2},  \label{eq:alpha_upate1}\\
    \mathcal{U}_{\sigma}  = \alpha  \mathcal{U}_{\sigma} + (1-\alpha ) \Sigma_{\sigma}^{2}, \label{eq:alpha_upate2}
\end{gather}
where $\mathcal{U}_{\mu} \in \mathbb{R}^{C}$ and $\mathcal{U}_{\sigma} \in \mathbb{R}^{C}$ 
denote the variance of global probabilistic distributions \textit{w.r.t.} $\mu(z)$ and $\sigma(z)$, respectively.
The values of $\mathcal{U}_{\mu}$ and $\mathcal{U}_{\sigma}$ are set as zeros at the beginning of the training, and become steady 
along the training. $\alpha$ is the momentum hyper-parameter to update $\mathcal{U}_{\mu}$ and $\mathcal{U}_{\sigma}$.

We sample style perturbations (the mean and standard deviation of new styles) from $\mathcal{N} (\mu, \mathcal{U}_{\mu})$ and $\mathcal{N} (\sigma, \mathcal{U}_{\sigma})$ using the re-parameterization trick \cite{kingma2013auto}:
\begin{gather}
    \beta(z) = \mu(z) + \xi_\mu \sqrt{\mathcal{U}_\mu},\\
     \gamma(z) = \sigma(z) + \xi_\sigma   \sqrt{\mathcal{U}_\sigma},
\end{gather}
where $\xi_* \sim  \mathcal{N} (0 ,1)$ are random variables drawing from a standard normal distribution.
Subsequently, we augment intermediate features $z$ by replacing the original style statistics in AdaIN \cite{huang2017adain} with the sampled style perturbations:
\begin{equation}
        z^*  = \gamma(z)(\frac{z-\mu (z)}{\sigma (z)} ) + \beta(z). \label{eq:adain}
\end{equation}

Here $z^*$ is the augmented intermediate features from $z$. 
Note that we update the global uncertainty statistics with both original samples and background-suppressed samples.

\begin{figure*}[htbp]  
  \centering 

  \includegraphics[width=0.9\linewidth]{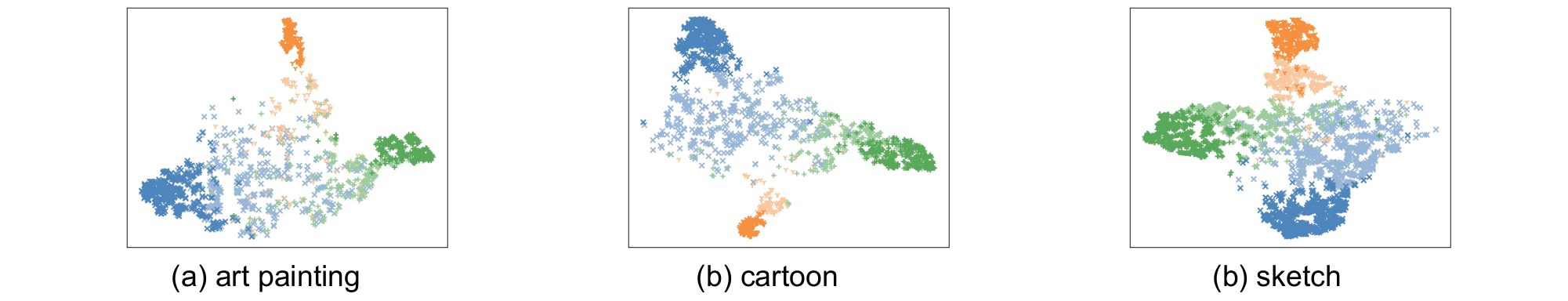}
\vspace{-0.1in}
\caption{The \textbf{t-SNE visualization} of the encoded feature distribution of original images and edge maps. The original image samples are represented in deep colors, while the corresponding edge maps are depicted in light colors. The samples are drawn from the `dog', `giraffe', and `guitar' categories of the three domains on PACS dataset.
}
\label{fig:edge_ood}
\end{figure*}

\subsubsection{Knowledge Distillation}\label{sec:knowledge_distill}
After obtaining the style-augmented samples with and without background, we perform a knowledge distillation to encourage the encoder to learn consistent representations regardless of the content and style perturbations. In particular, we take the background-suppressed augmentation as the distillation target since the removal of a background can restrict the spurious variable caused by an irrelevant context, and consequently reduce the intra-class variance. 
Formally, we denote the style-augmented features outputted by the encoder with and without background suppression as $o_{BS}\in \mathbb{R}^{C' \times H' \times W'}$ and $o \in \mathbb{R}^{C' \times H' \times W'}$, respectively. We distill consistent knowledge from $o_{BS}$ by computing the Kullback-Leibler (KL) divergence \cite{hershey2007kl} between $o$ and $o_{BS}$:
\begin{equation}
    \mathcal{L}_{kd} = - \frac{1}{H'W'} \sum_{C'} \sum_{H'} \sum_{W'} \delta (o_{BS}/\tau) \log \biggl\{\frac{ \delta(o/\tau)}{  \delta( o_{BS}/\tau)}\biggr\},
    \label{eq:loss_kd}
\end{equation}
where $\delta$ is the softmax function over the feature channel $C'$, and $\tau$ is the temperature hyperparameter.

\subsection{Boundary Growth}\label{sec:boundary_growth}
Our proposed domain expansion can help the model learn domain-invariant representations, however, it does not specifically take care of the open-set recognition problem in OS-SDG. Inspired by the recent success in addressing open-set identification by using the one-vs-all classifier for each class \cite{jang2022collective_one-vs-rest_OVA, saito2021ovanet, saito2021openmatch_OVA, zhu2022crossmatch_SDG}, we adopt multiple binary classifiers in this paper.  
%
Specifically, let $\omega_b = \left \{ (\omega_{b+}^{i}, \omega_{b-}^{i}) \right \} _{i=1}^{C^s} $ denote the multi-binary classifiers, where $\omega_{b+}^{i}$ and $\omega_{b-}^{i}$ are the positive and negative classifier of $i$-th class, respectively. The predictions (after softmax) of $\omega_{b+}^i$ and $\omega_{b-}^i$ are denoted as $p^i(y=1|x)$ and $p^i(y=0|x)$, where $p^i(y=1|x) + p^i(y=0|x) = 1$. $p^i(y=1|x)$ and $p^i(y=0|x)$ reflect the probability of an input $x$ belonging to the class $i$ and all other classes, respectively. For an image $x^s$ with label $y^s$, $x^s$ is the positive sample for $\omega_{b+}^{y^s}$ while might be the negative sample for other binary classifiers $\{\omega_{b-}^{j}\}_{j\ne y^s}^{C^s}$ based on the hard negative classifier
sampling strategy \cite{saito2021ovanet}. Each binary classifier is optimized by the one-vs-all (OVA) loss:
\begin{equation}
    \mathcal{L} _{ova}  = -\log(p^{y^s}(y = 1|x^s)) -\min_{ i\ne y^s} \log(p^i(y = 0|x^s)).
    \label{eq:loss_ova}
\end{equation}

The optimization of all multi-binary classifiers is inherently equivalent to extending the boundaries across all $|C^s|$ classes, and the efficacy of the boundary learning is heavily reliant on the selection of effective samples, \eg~hard negative samples. 

In view of this, we propose a novel classifier sampling strategy that incorporates edge maps as positive and negative samples to optimize the multi-binary classifiers.
Since the edge map of an image shares the same semantic information but different characteristics with the corresponding image, it can be considered as a new modality or view of the original image. We also observe that the feature distribution of edge maps within one class is shifted away from the original distribution of that class, as demonstrated in Figure \ref{fig:edge_ood}. This could be attributed to the fact that the removal of most appearance information from the edge map makes it become an out-of-distribution sample in the corresponding class, which is a better choice as the positive and hard negative sample for training multi-binary classifiers.

Let $\psi(.)$ denote the edge detector that extracts edge maps from images. Instead of the original $x^s$, we use $\psi(x^s)$ as the positive sample for $\omega_{b+}^{y^s}$. For $\{\omega_{b-}^{j}\}_{j\ne y^s}^{C^s}$, we use both $x^s$ and $\psi(x^s)$ as the hard negative samples. Thus, the optimization of each binary classifier in Eq.~\ref{eq:loss_ova} is modified as:

\begin{align}
    \mathcal{L}_{eova} = &-log(p^{y^s}(y =1|\psi(x^s)))   \notag\\
    &- \frac{1}{2}  \min_{ k\ne y^s} log(p^k(y = 0|\psi(x^s)))  \notag\\
    &-\frac{1}{2} \min_{ k\ne y^s} log(p^k(y = 0|x^s)).
    \label{eq:loss_eova}
\end{align}

 Since edge maps roughly represent out-of-distribution samples in the corresponding class, the optimization of the multi-binary classifiers with them can further grow the boundaries across known classes and thus improve the robustness of the model on unknown class identification.

\subsection{Training and Inference} \label{sec:train_and_infer}
During training, the model is optimized by the overall loss that combines the three losses:
\begin{eqnarray}
\mathcal{L}_{all} = \mathcal{L}_{ce} + \lambda_1 \mathcal{L}_{eova} + \lambda_2 \mathcal{L}_{kd}, \label{eq:loss_all}
\end{eqnarray}
where $\mathcal{L}_{ce}$ is the cross entropy loss computed between the outputs of multi-class classifier $\omega_{m}$ and the ground truths. $\lambda_1$ and $\lambda_2$ are hyper-parameters to control the contributions of the corresponding losses.

During inference, we obtain the prediction score (after softmax) from $\omega_{m}$ for each test image, and then we calculate the entropy of the prediction score. If the entropy value is smaller than a threshold, it is determined as the known class with the highest prediction score, otherwise, it is determined as an unknown class. To avoid threshold fine-tuning across datasets, we set the threshold to $\frac{1}{2} \log_{2}{\left | C^s \right | } $.


\section{Experiments}

\subsection{Experimental Settings}
\subsubsection{Datasets}
Following CrossMatch \cite{zhu2022crossmatch_SDG}, we evaluate our DEBUG on four cross-domain image classification datasets: 

\begin{itemize}[leftmargin=*]
\item \textbf{PACS} \cite{li2017pacs} contains 9,991 images of 7 categories collected from four domains: \textit{art painting}, \textit{cartoon}, \textit{photo}, and \textit{sketch}. 
Four classes (\ie~dog, elephant, giraffe, and guitar) are selected as the known classes, and the remaining classes (\ie~horse, house, and person) are treated as unknown.

\item \textbf{Office31} \cite{saenko2010office31} contains 4,652 images of 31 categories collected from three domains: \textit{Amazon}, \textit{DSLR} and \textit{webcam}. 
Ten classes are selected as known (\ie~back pack, bike, calculator, headphones, keyboard, laptop, monitor, mouse, mug, and projector), and another eleven classes are selected as unknown (\ie~ruler, punchers, stapler, scissors, trash can, tape dispenser, pen, phone, printer, ring binder, and speaker).

\item \textbf{OfficeHome} \cite{venkateswara2017officehome} contains 15,500 images of 65 categories from four domains: \textit{artistic}, \textit{clip art}, \textit{product}, and \textit{real-world}. 
The first 15 classes (\ie~alarm clock, backpack, battery, bed, bike, bottle, bucket, calculator, calendar, candles, chair, clipboards, computer, couch, and curtains) in alphabetic order are selected as known, and the rest 50 classes are treated as unknown. 

\item \textbf{DomainNet126} \cite{saito2019domainnet126} is a curated subset of the large-scale dataset DomainNet \cite{peng2019domainnet}. Compared to the original DomainNet, DomainNet126 is carefully constructed by removing certain domains and classes, thereby reducing the influence of noisy data. DomainNet126 has 126 classes and consists of four domains: \textit{real}, \textit{clipart}, \textit{painting}, \textit{sketch}. We select the first 63 classes in alphabetical order as the known classes and keep the remaining 63 classes as the unknown classes.

\begin{figure}[] 
\centering
\includegraphics[width=0.49\textwidth]{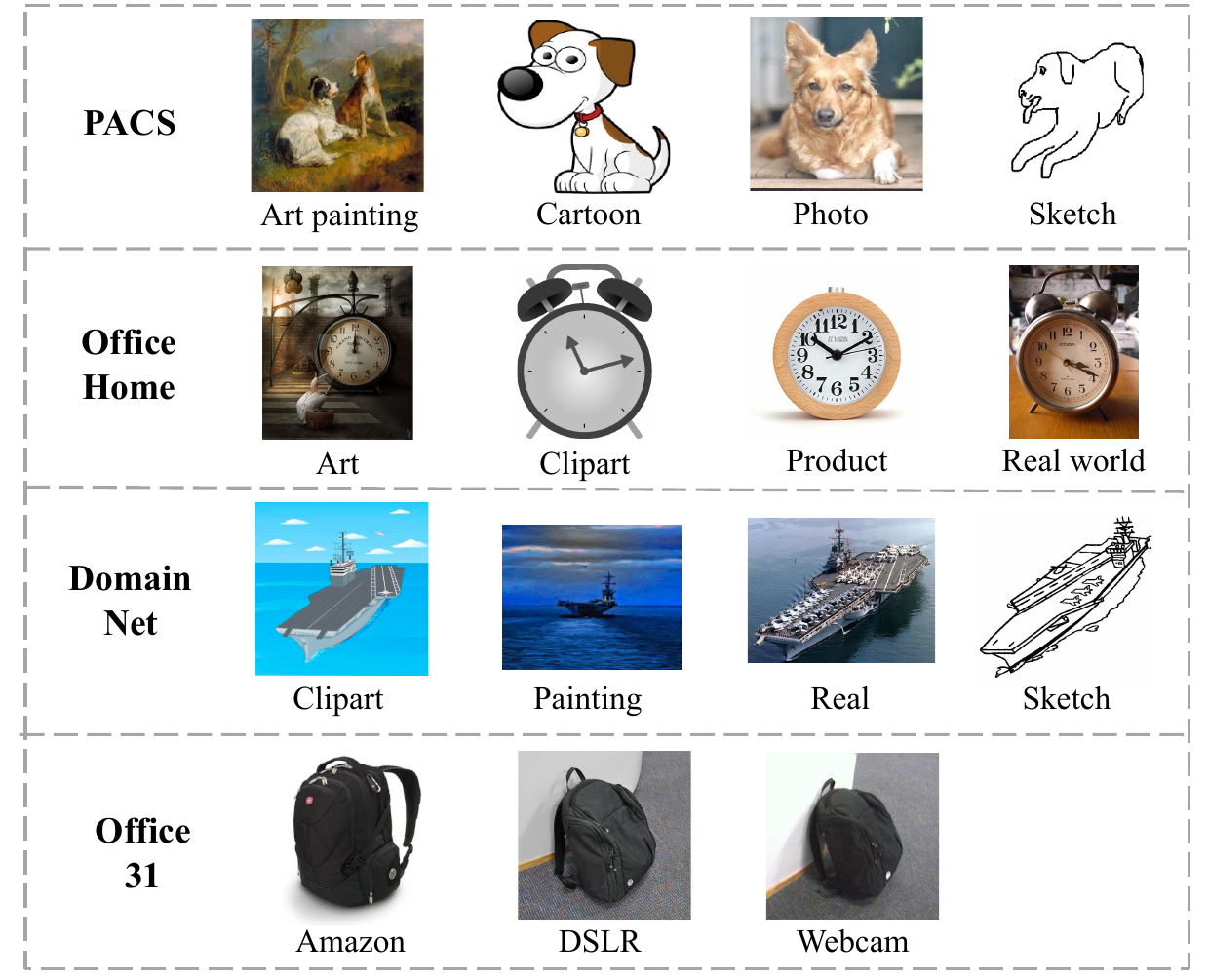} 
\caption{Examples across domains in different datasets.}
\label{fig:domain_example}
\end{figure}


\end{itemize}

Among these four datasets, PACS and DomainNet126 are considered more challenging datasets due to their greater domain diversity, as illustrated by the domain example comparisons in Figure \ref{fig:domain_example}.

We also follow evaluation settings as in CrossMatch.
Cross-validation is performed on the PACS, OfficeHome, and DomainNet126 datasets by iteratively selecting one domain as the source domain, and the rest of the domains as the targets. For the Office31 dataset, only Amazon is selected as the source domain, as DSLR and Webcam have relatively fewer samples.


\subsubsection{Implementation Details} \label{section: implementation}
We adopt the ImageNet pre-trained ResNet-18 \cite{he2015resnet} as the feature encoder. The multi-class classifier is one linear layer with output channel $|C^S|$, while each one-vs-all binary classifier is one linear layer with output channel 2.
We set the batch size to 32 and optimize the model using SGD with Nesterov momentum and a weight decay rate of 0.0005.
The learning rate is initially set to 0.001 and is scheduled by a step decay of 0.1 every 20 epochs. 
We set the temperature $\tau$ in Eq. \ref{eq:loss_kd} to 1.0, hyper-parameter $\lambda_1$ in Eq. \ref{eq:loss_all} as 0.5 on all the datasets, while we set $\lambda_2$ in Eq. \ref{eq:loss_all} as 1.0 on PACS, and 0.1 on Office31, OfficeHome, and DomainNet126. 
We use HED\cite{xie2015hed} to extract edge maps from the images.


\begin{table*} 
\centering
\caption{Performance comparison in terms of \textbf{overall accuracy} (\%) and \textbf{h-score} (\%) on \textbf{PACS} dataset.
}
\label{table:pacs_acc_hs}
\begin{tabularx}{0.9\textwidth}{l|CCCCCCCC|CC}
\toprule
\multirow{2}{*}{Method} & \multicolumn{2}{c}{Art}         & \multicolumn{2}{c}{Cartoon}     & \multicolumn{2}{c}{Sketch}      & \multicolumn{2}{c|}{Photo}      & \multicolumn{2}{c}{Average}     \\

\cmidrule(lr){2-3} \cmidrule(lr){4-5} \cmidrule(lr){6-7} \cmidrule(lr){8-9} \cmidrule(lr){10-11}
                        & acc            & hs             & acc            & hs             & acc            & hs             & acc            & hs             & acc            & hs             \\ 
                        
\midrule
ERM+CM                  & 53.52          & 44.90           & 57.60           & 48.31          & 38.53          & 30.43          & 42.52          & 41.60           & 50.54          & 41.31          \\
MEADA+CM               & 62.63          & 41.88          & 60.03          & 51.36          & 41.51          & 35.76          & 43.50           & 41.60           & 51.92          & 42.65          \\
ADA+CM                 & 64.26          & 42.40           & 60.41          & 51.81          & 42.48          & 35.18          & 43.97          & 42.76          & 52.78          & 43.04          \\
\midrule
DSU+OVA & 60.83 & 42.53 & \textbf{63.30}     & \textbf{61.38}     & 41.76 & 41.13 & 43.61 & 39.58     &52.37  & 46.15 \\
L2D+OVA & 61.65	    & \textbf{64.61}     & 49.68     & 53.56     & 47.50 	& 51.15    & 40.78     & 42.66     & 49.91     & 52.99 \\
\midrule
\textbf{DEBUG} (ours)                                              & \textbf{69.52} & 57.85      & 60.79     & 59.05  & 49.95 & 48.75 & \textbf{50.63} & \textbf{52.43} & \textbf{57.72} & \textbf{54.52} \\ 
\bottomrule
\end{tabularx}
\end{table*}

\begin{table*} 
\centering
\caption{Performance comparison in terms of \textbf{known} and \textbf{unknown class accuracy} (\%) on \textbf{PACS} dataset.}
\label{table:pacs_acc_k_u}
\begin{tabularx}{0.9\textwidth}{l|CCCCCCCC|CC}
\toprule
\multirow{2}{*}{Method} & \multicolumn{2}{c}{Art}         & \multicolumn{2}{c}{Cartoon}     & \multicolumn{2}{c}{Sketch}      & \multicolumn{2}{c|}{Photo}      & \multicolumn{2}{c}{Average}     \\

\cmidrule(lr){2-3} \cmidrule(lr){4-5} \cmidrule(lr){6-7} \cmidrule(lr){8-9} \cmidrule(lr){10-11}
                        & $\mathrm{acc_k}$                           & $\mathrm{acc_u}$                         & $\mathrm{acc_k}$                   & $\mathrm{acc_u}$                           & $\mathrm{acc_k}$                            & $\mathrm{acc_u}$                         & $\mathrm{acc_k}$                            & $\mathrm{acc_u}$                            & $\mathrm{acc_k}$        & $\mathrm{acc_u}$       \\ 
\midrule
ERM+CM                  & 68.66          & 44.56          & 62.65          & 43.18          & 41.01          & 33.16          & 39.91          & 54.21         & 52.96          & 44.53          \\
MEADA+CM                & 70.45          & 33.36          & 63.76          & 53.74          & 40.25          & 48.79          & 42.89          & 50.57         & 54.34          & 46.61          \\
ADA+CM                  & 72.93          & 40.12          & 64.39 & 49.06          & 44.98          & 40.85          & 43.27          & 52.53         & 56.40          & 45.64          \\
\midrule
DSU+OVA & 68.17     & 31.50     & \textbf{64.47}	    & 58.60   & 37.16   &60.17  & 44.95	    & 38.22    & 53.69  & 47.12  \\
L2D+OVA & 56.00     & \textbf{84.30}     & 45.69     & \textbf{65.63}     & 40.15     & \textbf{76.93}     & 33.08     & \textbf{71.56}     & 43.73     & \textbf{74.60} \\
\midrule
\textbf{DEBUG} (ours)                   & \textbf{74.75} & 48.63 & 61.66          & 57.28 & \textbf{45.35} & 68.32 & \textbf{48.78} & 58.05 & \textbf{57.63} & 58.07 \\ 
\bottomrule
\end{tabularx}
\end{table*}

\begin{table*} 
\centering
\caption{Performance comparison in terms of \textbf{overall accuracy} (\%) and \textbf{h-score} (\%) on \textbf{OfficeHome} dataset.}
\label{table:officehome_acc_hs}
\begin{tabularx}{0.9\textwidth}{l|CCCCCCCC|CC}
\toprule
\multirow{2}{*}{Method} & \multicolumn{2}{c}{Artistic}    & \multicolumn{2}{c}{Clip Art}    & \multicolumn{2}{c}{Product}     & \multicolumn{2}{c|}{Real-World} & \multicolumn{2}{c}{Average}     \\
\cmidrule(lr){2-3} \cmidrule(lr){4-5} \cmidrule(lr){6-7} \cmidrule(lr){8-9} \cmidrule(lr){10-11}
                        & acc            & hs             & acc            & hs             & acc            & hs             & acc            & hs             & acc            & hs             \\ 
\midrule
ERM+CM                  & 65.49          & 52.85          & \textbf{63.37} & 50.51          & 58.03          & 47.25          & 67.75          & 52.60          & \textbf{63.66}    & 50.80            \\
MEADA+CM                & 65.85          & 53.22          & 62.90          & 48.87          & 58.36          & 45.34          & 67.10          & 50.77          & 63.55    & 49.55            \\
ADA+CM                  & \textbf{66.30} & 46.68          & 62.64          & 49.31          & 58.72          & 47.47          & 66.82          & 50.47          & 63.62    & 48.48            \\
\midrule
DSU+OVA & 64.42 & 51.45     & 62.57     & \textbf{56.60}     & 53.61   & 50.12     & 59.43 & 45.25    & 60.01  &50.85 \\
L2D+OVA & 58.84     & \textbf{57.47}     & 59.20     & 41.30     & 53.85     & 52.25     & 59.38     & 45.99     & 57.83       & 49.28 \\
\midrule
\textbf{DEBUG} (ours)                & 63.16          & 57.15 & 61.28          & 55.90 & \textbf{59.90} & \textbf{53.45} & \textbf{67.80} & \textbf{59.38} & 63.03    & \textbf{56.47}   \\ 
\bottomrule
\end{tabularx}
\end{table*}

\begin{table*}
\centering
\caption{Performance comparison in terms of \textbf{known} and \textbf{unknown class accuracy} (\%) on \textbf{OfficeHome} dataset.}
\label{table:officehome_acc_k_u}
\begin{tabularx}{0.9\textwidth}{l|CCCCCCCC|CC}
\toprule
\multirow{2}{*}{Method} & \multicolumn{2}{c}{Art}         & \multicolumn{2}{c}{Clipart}     & \multicolumn{2}{c}{Product}      & \multicolumn{2}{c|}{Real}      & \multicolumn{2}{c}{Average}     \\

\cmidrule(lr){2-3} \cmidrule(lr){4-5} \cmidrule(lr){6-7} \cmidrule(lr){8-9} \cmidrule(lr){10-11}
                        & $\mathrm{acc_k}$                           & $\mathrm{acc_u}$                         & $\mathrm{acc_k}$                   & $\mathrm{acc_u}$                           & $\mathrm{acc_k}$                            & $\mathrm{acc_u}$                         & $\mathrm{acc_k}$                            & $\mathrm{acc_u}$                            & $\mathrm{acc_k}$        & $\mathrm{acc_u}$       \\ 
\midrule
ERM+CM                  & 66.48          & 48.57          & \textbf{64.80} & 41.95          & 59.17          & 40.94         & \textbf{69.36} & 43.69          & 64.95          & 43.79          \\
MEADA+CM                & 66.63          & 45.28          & 64.43          & 37.84          & 59.74          & 37.71         & 68.82          & 41.28          & 64.90          & 40.53          \\
ADA+CM                  & \textbf{67.53} & 39.59          & 64.10          & 40.67          & 59.92          & 40.72         & 68.53          & 40.79          & \textbf{65.02} & 40.44          \\
\midrule
DSU+OVA & 65.86     & 42.75    & 63.27  & 51.98     & 53.88    &49.53    
 & 60.96    &36.60  & 60.99 &45.22 \\
 L2D+OVA & 58.92     & \textbf{57.53}   & 61.03       & 31.86      & 53.90    & \textbf{53.20}     & 60.81     & 37.90     & 58.68       & 45.15 \\
 \midrule
\textbf{DEBUG} (ours)                    & 63.80           & 53.47        & 61.86          & \textbf{52.58} & \textbf{60.57} & 49.80 & 68.76          & \textbf{53.25} & 63.75          & \textbf{52.28} \\
\bottomrule
\end{tabularx}
\end{table*}

\begin{table*} 
\centering
\caption{Performance comparison in terms of \textbf{overall accuracy} (\%) and \textbf{h-score} (\%) on \textbf{DomainNet126} dataset.} 
\label{table:domainnet126_acc_hs}
\begin{tabularx}{0.9\textwidth}{l|CCCCCCCC|CC}
\toprule
\multirow{2}{*}{Method} & \multicolumn{2}{c}{Clipart}         & \multicolumn{2}{c}{Painting}     & \multicolumn{2}{c}{Real}      & \multicolumn{2}{c|}{Sketch}      & \multicolumn{2}{c}{Average}     \\

\cmidrule(lr){2-3} \cmidrule(lr){4-5} \cmidrule(lr){6-7} \cmidrule(lr){8-9} \cmidrule(lr){10-11}
                        & acc            & hs             & acc            & hs             & acc            & hs             & acc            & hs             & acc            & hs             \\ 
                        
\midrule
DSU+OVA     & 46.68     & 46.04     & 52.06     & 45.82     & 50.76     & 41.44     & 48.42     & 43.3      & 49.48     & 44.15 \\
L2D+OVA    & 47.72  & 45.16  & 51.82  & 46.77  & \textbf{52.39}  & 43.22  & \textbf{50.49}  & 47.68  & 50.61  & 45.71 \\ \midrule
\textbf{DEBUG} (ours)                                              & \textbf{49.12} & \textbf{46.56}      & \textbf{55.48}  & \textbf{47.16} & 51.74 & \textbf{43.75} & 50.36 & \textbf{48.19} & \textbf{51.67}  &\textbf{46.41}\\ 
\bottomrule
\end{tabularx}
\end{table*}

\begin{table*}
\centering
\caption{Performance comparison in terms of \textbf{known} and \textbf{unknown class accuracy} (\%) on \textbf{DomainNet126} dataset.}
\label{table:domainnet126_acc_k_u}
\begin{tabularx}{0.9\textwidth}{l|CCCCCCCC|CC}
\toprule
\multirow{2}{*}{Method} & \multicolumn{2}{c}{Clipart}         & \multicolumn{2}{c}{Painting}     & \multicolumn{2}{c}{Real}      & \multicolumn{2}{c|}{Sketch}      & \multicolumn{2}{c}{Average}     \\

\cmidrule(lr){2-3} \cmidrule(lr){4-5} \cmidrule(lr){6-7} \cmidrule(lr){8-9} \cmidrule(lr){10-11}
                        & $\mathrm{acc_k}$             & $\mathrm{acc_u}$              & $\mathrm{acc_k}$             & $\mathrm{acc_u}$              & $\mathrm{acc_k}$             & $\mathrm{acc_u}$              & $\mathrm{acc_k}$             & $\mathrm{acc_u}$              & $\mathrm{acc_k}$             & $\mathrm{acc_u}$              \\ 
                        
\midrule
DSU+OVA     & 46.69     & \textbf{46.04}     & 52.19     & 43.35     & 51.01     & 35.32     & 48.56     & 39.58      & 49.61     & 41.07 \\
L2D+OVA    & 47.80  & 43.08  & 51.95  & \textbf{43.80}  & \textbf{52.64}  & 36.93  & \textbf{50.57}  & 45.70  & 50.74  & 42.37 \\
 \midrule
\textbf{DEBUG} (ours)                                              & \textbf{49.19} & 44.60      & \textbf{55.68}  & 42.27 & 51.96 & \textbf{38.13} & 50.42 & \textbf{46.69} & \textbf{51.81}  &\textbf{42.92}\\ 
\bottomrule
\end{tabularx}
\end{table*}

\begin{table} 
\centering
\caption{Performance comparison on \textbf{Office31} dataset. 
}
\label{table:office31_acc_hs}
\scalebox{1.1}{
\begin{tabular}{p{3cm}|ll|ll}
\toprule
\multicolumn{1}{l|}{Method}   & \multicolumn{1}{c}{$\mathrm{acc_k}$} & \multicolumn{1}{c|}{$\mathrm{acc_u}$} & \multicolumn{1}{c}{$\mathrm{acc}$} & \multicolumn{1}{c}{$\mathrm{hs}$}  \\ 
\midrule
ERM+CM                    & 82.37          & 37.60          & 78.30          & 51.14          \\
MEADA+CM                  & 82.77          & 41.08          & \textbf{78.98}          & 54.69          \\
ADA+CM                    & \textbf{83.02}          & 34.51          & 78.61          & 48.50          \\
\midrule
DSU+OVA     & 75.75   & 58.83     & 74.21     & 65.99   \\
L2D+OVA     & 73.97   & 60.90     & 72.78     & 66.41   \\
\midrule
\textbf{DEBUG}                      & 75.04         & \textbf{65.28} & 74.15         & \textbf{69.40} \\
\bottomrule
\end{tabular}}
\end{table}

\subsubsection{Baselines} \label{sec:exp_baselines}
We compare our DEBUG with three methods introduced in \cite{zhu2022crossmatch_SDG}, which integrate CrossMatch with Empirical Risk Minimization (ERM) \cite{koltchinskii2011erm}, and two adversarial generation based methods, \ie~Adversarial Data Augmentation (ADA) \cite{volpi2018ada_SDG} and Maximum-Entropy Adversarial Data Augmentation (MEADA) \cite{zhao2020meada_SDG}, resulting in ERM+CM, ADA+CM, and MEADA+CM. 
We also extend two style augmentation methods (\ie~L2D \cite{wang2021learning2diversify_SDG} and DSU \cite{li2022uncertainty}) with the typical open-set recognition method OVANet \cite{saito2021ovanet} (\ie~the one-vs-all loss in Eq. \ref{eq:loss_ova}) as our baselines, denoted as L2D+OVA and DSU+OVA. 


\subsubsection{Evaluation Metrics}
We use the same evaluation metrics as CrossMatch, \ie~known class accuracy ($acc_k$), unknown class accuracy ($acc_u$), overall mean accuracy ($acc$), and h-score ($hs$) which is computed as $hs=\frac{2 \ast acc_k \ast acc_u}{acc_k + acc_u}$.
Note that all the unknown classes are regarded as one class. 
It is worth highlighting that the harmonic mean $hs$ provides a more precise measure than $acc$ as it accounts for the discrepancies in accuracy between the known and unknown classes.

\subsection{Experimental Results}
By conducting a comprehensive performance comparison with the baseline methods across four datasets, our steady performance improvements over the baselines in terms of the h-score validate the effectiveness of our proposed DEBUG in the context of OS-SDG. We will now proceed to analyze the experimental results for each dataset individually. 

\subsubsection{Results on PACS} 
Table \ref{table:pacs_acc_hs} and Table \ref{table:pacs_acc_k_u} present comparisons of DEBUG with various baselines in terms of \{$acc$, $hs$, $acc_k$, $acc_u$\} on the PACS dataset. 
Table \ref{table:pacs_acc_hs} clearly demonstrates the significant superiority of our DEBUG method over the previous state-of-the-art baseline ADA+CM across all the domains. Our method achieves an impressive 9\% increase in overall accuracy, and a remarkable 26\% increase in h-score. These results underscore the effectiveness of our proposed domain expansion and boundary growth techniques within known classes when compared to adversarial data augmentation. 
Among the two style augmentation methods, L2D+OVA demonstrates relatively high performance in $hs$, particularly excelling in recognizing unknown classes, as indicated by the $acc_u$ scores in Table \ref{table:pacs_acc_k_u}. However, it does exhibit a noticeable drop in accuracy for known classes, which could lead to a higher likelihood of misclassifying close-set samples as unknown. 
In contrast, our DEBUG manages to strike a balance in the model's ability to recognize both the known and unknown classes, as indicated by the h-score ($hs$), which offers a more accurate evaluation, taking into account the performance imbalance between the known and unknown classes.


\subsubsection{Results on Office31} 
Table \ref{table:office31_acc_hs} reports the comparison results for Office31. 
As illustrated in Figure \ref{fig:domain_example}, Office31 exhibits relatively small domain shifts between the cross-domain samples, with similarities in style and background. In such a scenario, the three methods introduced in \cite{zhu2022crossmatch_SDG} struggle to distinguish the unknown classes, often confusing them with the known classes. In contrast, our DEBUG method excels in accurately recognizing the unknown classes while maintaining a strong performance on the known classes, ultimately achieving the best h-score performance.

\subsubsection{Results on OfficeHome}
OfficeHome is more challenging than Office31 since it contains more unknown classes and larger domain shifts. Table \ref{table:officehome_acc_hs} and Table \ref{table:officehome_acc_k_u} list the comparison results between DEBUG and the baselines on OfficeHome with \{$acc$, $hs$, $acc_k$, $acc_u$\} metrics. 
Table \ref{table:officehome_acc_k_u} reveals that the adversarial data augmentation-based methods exhibit superior performance in recognizing the known classes but lower performance in identifying the unknown classes. On the other hand, the two style augmentation methods show the opposite trend. Our DEBUG method effectively mitigates the shortcomings of both techniques, achieving strong performances in recognizing both the known and unknown classes. Consequently, it achieves a significant improvement in h-score compared to the baselines, as shown in Table \ref{table:officehome_acc_hs}.


\subsubsection{Results on DomainNet126} 
DomainNet126 is the most challenging among the four datasets due to its domain diversity, an extensive array of categories, and a substantial volume of samples per category. When experimenting on DomainNet126, we set the batch size to 96 and the learning rate to 0.003, considering the substantial volume of samples.

Table \ref{table:domainnet126_acc_hs} and Table \ref{table:domainnet126_acc_k_u} present comparisons of our DEBUG with the two style augmentation methods in terms of \{$acc$, $hs$, $acc_k$, $acc_u$\} on the DomainNet126 dataset. Since CrossMatch \cite{zhu2022crossmatch_SDG} did not provide an evaluation on the DomainNet126 dataset and also did not share their code with sufficient implementation details for re-implementation, we have excluded the comparison with the three baselines from \cite{zhu2022crossmatch_SDG} on this dataset.
As depicted in the tables, our proposed DEBUG approach consistently outperforms the two baseline methods, showcasing the superior performance of our approach in both domain generalization and open-set recognition.

\begin{table*}[t] 
\centering
\caption{\textbf{The effects of our proposed components on PACS}. BS represents background suppression, GPSA denotes global probabilistic-based style augmentation, KD represents knowledge distillation, and EOVA indicates the use of edge maps in computing one-vs-all (OVA) loss in Eq.~\ref{eq:loss_eova}.}
\label{table:ablation}
\begin{tabularx}{0.8\textwidth}{c|CCC|c|CCCC}
\toprule
 \multirow{2}{*}{Method}         & \multicolumn{3}{c|}{Domain Expansion}                & Boundary Growth     &    \multirow{2}{*}{$\mathrm{acc_k}$} &    \multirow{2}{*}{$\mathrm{acc_u}$}      &    \multirow{2}{*}{acc}             &          \multirow{2}{*}{hs}       \\
 \cmidrule(lr){2-4}  \cmidrule(lr){5-5}
     & BS                        & GPSA                      & KD                        & EOVA                      &             &              \\
\midrule
DSU+OVA   &                           &                           &               &            & 53.69        & 47.12        & 52.37        & 46.15 \\
\midrule
variant 1 & &  & \checkmark  & & 54.37&	49.28	&53.35	& 48.35
 \\
variant 2 & &  \checkmark  & & &53.805 & 50.87	& 53.21 &	47.27
 \\
variant 3 &    \checkmark             &                           & \checkmark    &            & 56.41        & 52.07        & 55.54        & 51.93 \\
variant 4 &                           & \checkmark                & \checkmark    &            & 55.08        & 49.19        & 53.91        & 47.11 \\
variant 5 & \checkmark                & \checkmark                & \checkmark    &            & \textbf{59.9}   & 49.58        & \textbf{57.84}        & 52.92 \\
variant 6 &                           &                           &                           & \checkmark   & 53.42   & 52.05   & 53.15   & 50.75 \\\midrule
variant 7 & \checkmark                &                           & \checkmark                & \checkmark   & 55.85   & 54.12   & 55.51   & 52.93 \\
variant 8 &                           & \checkmark                & \checkmark                & \checkmark   & 56.96   & 51.28   & 55.82   & 52.20  \\
\midrule
\textbf{DEBUG}     & \checkmark                & \checkmark                & \checkmark                & \checkmark          & 57.63    & \textbf{58.07}      & 57.72 & \textbf{54.52} \\
\bottomrule
\vspace{-0.2in}
\end{tabularx}
\end{table*}

\begin{table}[]
\centering
\caption{\textbf{Ablation study of the Domain Expansion and Boundary Growth on Office31, OfficeHome and DomainNet126.} “DE” denotes the Domain Expansion module, while “BG” represents the Boundary Growth module.}
\label{tbl:abla_on_office31_officehome}
\begin{tabular}{lllllll}
\toprule
Dataset    & DE & BG & $\mathrm{acc_k}$ & $\mathrm{acc_u}$ & acc   & hs    \\
\midrule
\multirow{3}{*}{Office31}     &    &    & \textbf{85.10}  & 27.04  & 79.82 & 40.69 \\
           & \checkmark  & & 83.23 & 51.6 & \textbf{80.36} & 63.62    \\
           &    & \checkmark  & 68.31  & 61.93  & 67.73 & 64.6  \\
           & \checkmark  & \checkmark  & 75.04  & \textbf{65.28}  & 74.15 & \textbf{69.4}  \\ 
\midrule
\multirow{3}{*}{OfficeHome} &    &    & 66.21  & 41.62  & 64.67 & 49.71 \\
           & \checkmark  &    & \textbf{67.58}  & 47.02  & \textbf{66.3}  & 54.11 \\
           &    & \checkmark  & 60.63  & 52.07  & 60.09 & 54.97 \\
           & \checkmark  & \checkmark  & 63.75  & \textbf{52.28}  & 63.03 & \textbf{56.47} \\
\midrule
\multirow{3}{*}{DomainNet126} &    &    & 46.56  & 41.36  & 46.48 & 42.07 \\
           & \checkmark  &    & \textbf{53.56}  & 32.28  & \textbf{53.22}  & 39.55 \\
           &    & \checkmark  & 32.29  & \textbf{68.12}  & 32.85 & 42.43 \\
           & \checkmark  & \checkmark  & 51.81  & 42.92  & 51.67 & \textbf{46.41} \\

\bottomrule
\end{tabular}
\end{table}

\subsection{Additional Analysis}
\subsubsection{Ablation Study}
We conduct the ablation study on \textbf{PACS} to validate the effectiveness of each component in our proposed method. Table \ref{table:ablation} presents a comparative analysis of our approach with the closest baseline (\ie~DSU+OVA),
six variants of our model, and our full model. 
 
\setlength\parindent{0pt}
\textbf{Domain Expansion.} When comparing variants 3 and 4 to the baseline, we observe that both variants exhibit notable performance improvements. 
These enhancements are attributed to the incorporation of knowledge distillation, which enforces consistency regularization, whether through \textit{content} (as in variant 3 with BS) or \textit{style} (as in variant 4 with GPSA) perturbations. This regularization effectively reduces intra-class variance, subsequently enhancing the model's robustness.
The additional improvement in performance on $acc_k$ observed when comparing variant 5 to the baseline underscores that our domain expansion approach, \ie~applying knowledge distillation in the presence of both content and style perturbations, can substantially enhance the model's ability to generalize to cross-domain samples.
The performance improvements observed when comparing our DEBUG method to variant 7, where GPSA is replaced with DSU, highlight the effectiveness of our global probabilistic distribution over the batch-wise probabilistic distribution in DSU for open-set single-source domain generalization (as discussed in Section~\ref{sec:gp_style_aug}). Similarly, the performance enhancements seen in the comparison between our DEBUG method and variant 8, underscore the effectiveness of our background suppression scheme. 

\textbf{Boundary Growth.}
Comparing variant 6 with the baseline, the performance improvement on $acc_u$ clearly shows that boundary growth can increase the model capacity in open-set recognition.
In variant 5, our full domain expansion mechanism significantly enhances the recognition of the known classes, yet it only shows marginal enhancements in relation to the unknown classes. In our final DEBUG method, the incorporation of EOVA distinctly promotes the growth of the boundary that separates the known and unknown classes, ultimately resulting in the best performance.

\textbf{Ablation of Key Components on Office31, OfficeHome and DomainNet126 Datasets.}
We further conduct ablation studies on the OfficeHome, Office31, and DomainNet126 datasets, focusing on the overall structure of the Domain Expansion and Boundary Growth modules. The results are presented in Table~\ref{tbl:abla_on_office31_officehome}, where the ERM (please refer to Section~\ref{sec:exp_baselines} for more details) setting serves as the baseline. As observed, the results align with those obtained from the PACS dataset, the Domain Expansion module improves the overall accuracy, while the Boundary Growth module is effective in unknown class recognition. The model performs best on the h-score when these two components are combined. These results further show the effectiveness of our proposed DEBUG approach.

\begin{table}[]
\centering
\caption{\textbf{Ablation study of background suppression (BS) on PACS.} DEBUG$\mathrm{^{\dagger}}$ removes BS, DEBUG$\mathrm{^{\ddagger}}$ uses \cite{hou2007sailencyDetect} to achieve BS. The term “PRP” stands for the average percentage of removed pixels.}
\label{table:ablation_BS}
\scalebox{1}{
\begin{tabular}{lcccc}
\toprule
&  & DEBUG$\mathrm{^{\dagger}}$ & DEBUG$\mathrm{^{\ddagger}}$ & DEBUG \\
\midrule
\multirow{3}{*}{PACS} & PRP & 0\% & 47.4\% & 62.86\%  \\
& acc    & 55.82 & 56.58 & \textbf{57.72} \\
& hs     & 52.20 & 53.07 & \textbf{54.52} \\

\midrule
\multirow{3}{*}{Office31} & PRP & 0\% & 36.18\% & 78.03\%  \\
& acc    & 73.33 & 71.35 & \textbf{74.15} \\
& hs     & 68.8 & 69.27 & \textbf{69.4} \\
\midrule
\multirow{3}{*}{OfficeHome} & PRP & 0\% & 47.94\% & 76.57\%  \\
& acc    &  \textbf{65.07} & 64.79 & 63.03 \\
& hs     & 53.94 & 54.79 & \textbf{56.47} \\
\bottomrule
\end{tabular}}
\end{table}

\begin{table}[]
\centering
\caption{\textbf{Hyperparameter Selection on PACS.}}

\label{table:ablation_alpha}
\scalebox{1}{
\begin{tabular}{cccccc}
\toprule
  & $\alpha$=0.9 & $\alpha$=0.8 & $\alpha$=0.7 & $\alpha$=0.6 & $\alpha$=0.5 \\
hs & 53.24    & \textbf{54.52} & 53.43 &  53.12 & 52.44\\
\midrule
  & $\lambda_1$=0.1 & $\lambda_1$=0.5 & $\lambda_1$=1.0 & $\lambda_1$=1.5 & $\lambda_1$=2.0 \\
hs &  52.44  & 53.74 &\textbf{54.52}&  54.39 & 54.1\\
\midrule
  &  & $\lambda_2$=0.5 & $\lambda_2$=1.0 & $\lambda_2$=1.5 & $\lambda_2$=2.0 \\
hs &     & 53.76 &\textbf{54.52}&  53.03 & 52.87\\
\bottomrule
\end{tabular}}
\end{table}



\subsubsection{Hyperparameter Selection}
We perform tests using various values of the hyperparameters, including $\alpha$, $\lambda_1$, and $\lambda_2$. For detailed information about $\alpha$, please refer to Equation~\ref{eq:alpha_upate1}; for $\lambda_1$ and $\lambda_2$, please refer to Equation~\ref{eq:loss_all}. The results on the PACS dataset can be found in Table \ref{table:ablation_alpha}. To simplify the process, we fix $\alpha$ at 0.8 and $\lambda_1$ at 1.0, and only vary the hyperparameter $\lambda_2$ for the other three datasets for hyperparameter selection.

\vspace{0.1in}
\subsubsection{Analysis on Background Suppression}
To verify the impact of the employed open-vocabulary segmentation method \ie~DenseCLIP on the efficacy of our method, we compare our DEBUG with two additional baselines: 1) DEBUG$\mathrm{^{\dagger}}$ that does not incorporate any background suppression, and 2) DEBUG$\mathrm{^{\ddagger}}$ that substitutes the class-aware DenseCLIP \cite{zhou2021denseclip} with a class-agnostic saliency detection method \cite{hou2007sailencyDetect} to produce foreground masks. 
Table \ref{table:ablation_BS} presents the comparison results across the PACS, Office31, and OfficeHome datasets. As can be seen from the table, while the incorporation of background suppression facilitates the improved performance of our method (DEBUG$\mathrm{^{\dagger}}$ $v.s.$ DEBUG), our method exhibits a degree of robustness to the specific choice of segmentation methods (DEBUG$\mathrm{^{\ddagger}}$ $v.s.$ DEBUG).

\begin{table}[]
\centering
\caption{
\textbf{Evaluation of two substitutes for edge maps in boundary expansion on the PACS dataset}. Comprehensive descriptions of these alternatives are provided in the text.
}
\label{table:ablation_EOVA}
\scalebox{1}{
\begin{tabular}{c|cccc}
\toprule
\multicolumn{1}{c|}{ } & \begin{tabular}[c]{@{}c@{}}original\\ (OVA)\end{tabular} & \begin{tabular}[c]{@{}c@{}}weak\\ augmentation\end{tabular} & \begin{tabular}[c]{@{}c@{}}strong\\ augmentation\end{tabular} & \begin{tabular}[c]{@{}c@{}}edge map\\ (EOVA)\end{tabular}      \\
\midrule
$\mathrm{acc_k}$ & \textbf{60.13}   & 56.39   & 57.65   & 57.63     \\
$\mathrm{acc_u}$ & 47.76   &  55.78  & 55.10   & \textbf{58.07}     \\
acc                        & 57.66  & 56.27    & 57.14               & \textbf{57.72} \\
hs                         & 51.51  & 53.35    & 53.74               & \textbf{54.52} \\
\bottomrule
\end{tabular}
}
\end{table}


\begin{table}[]
\centering
\caption{
\textbf{Evaluation of two substitutes for edge maps in boundary expansion on the \textit{constituted} PACS dataset}, by excluding the sketch domain.
}
\label{table:no_sketch}
\resizebox{1\columnwidth}{!}{
\begin{tabular}{c|cccc}
\toprule
\multicolumn{1}{c|}{ } & \begin{tabular}[c]{@{}c@{}}original\\ (OVA)\end{tabular} & \begin{tabular}[c]{@{}c@{}}weak\\ augmentation\end{tabular} & \begin{tabular}[c]{@{}c@{}}strong\\ augmentation\end{tabular} & \begin{tabular}[c]{@{}c@{}}edge map\\ (EOVA)\end{tabular}      \\
\midrule
$\mathrm{acc_k}$ & 66.14   & 66.74  & \textbf{68.48}  & 66.33     \\
$\mathrm{acc_u}$ & 51.94   & 52.70  & 53.66   & \textbf{58.44}     \\
acc                        & 63.30  & 63.93             & \textbf{65.52}               & 64.75 \\
hs                         & 57.74  & 58.44             & 59.77               & \textbf{61.42} \\
\bottomrule
\end{tabular}
}
\end{table}

\begin{table}[]
\centering
\caption{Results of Using Different Feature Encoders on PACS.}
\label{tbl:model_arch}
\begin{tabular}{llllll}
\toprule
Backbone &  Method & $\mathrm{acc_k}$ & $\mathrm{acc_u}$ & acc & hs \\
\midrule
\multirow{2}{*}{ResNet-18} & DSU+OVA  &  53.69  &  47.12  &  52.37  &  46.15  \\
   &  DEBUG & 57.63  & 58.07  & 57.72 & 54.52 \\
\midrule
\multirow{2}{*}{ResNet-50} & DSU+OVA   & 58.55 &  54.64  & 57.77 &  54.19\\
         &  DEBUG  & 63.47  & 58.59  & 62.50       & 58.27   \\
\midrule
\multirow{2}{*}{ResNet-101} & DSU+OVA  & 57.44 &    60.54 &    58.06 &    54.23    \\
          &  DEBUG & 67.26 &  54.91 &  64.79 &  59.04   \\
\bottomrule
\end{tabular}
\end{table}

\vspace{0.1in}
\subsubsection{Analysis on Edge Maps}
In Table \ref{table:ablation_EOVA}, we evaluate the impact of the edge maps by replacing them with alternative design choices for the selection of positive and hard negative samples during the optimization of multi-binary classifiers. 
Two alternatives are: 1) \textit{weak augmentation}, where we apply random crop and random horizontal flip to each image to produce a weak augmentation, and 2) \textit{strong augmentation}, where we additionally apply random grayscale and Gaussian blurring to produce a strong augmentation. 
In Table \ref{table:ablation_EOVA}, we observe that utilizing augmented images to select positive or negative samples yields an improved open-set recognition performance compared to using the original images. Furthermore, the extent of improvement increases with the augmentation strength. We posit that augmentations effectively shift the original distribution, generating out-of-distribution (OOD) samples. These OOD samples can be advantageous for training multi-binary classifiers, promoting boundary growth between the known and unknown classes. Notably, among the options for generating OOD samples, using edge maps as a modality is a better choice than using weak or strong augmentations. This is because edge maps convey less appearance-specific information and can, therefore, generate more effective OOD samples, resulting in superior performances.  

\setlength\parindent{15pt}
To eliminate the potential effect of edge maps when the target domain on PACS is the sketch domain, we remove the sketch domain and keep the remaining three to constitute a new dataset on PACS. We further conduct experiments on this constituted PACS dataset and report the results in Table \ref{table:no_sketch}. As can be seen from the table, our EOVA method still achieves the best performance on the h-score compared to the other alternatives. 
We notice that using the strong augmentation outperforms the edge map in identifying known classes ($\mathrm{acc_k}$); this is possible because the strong augmentation may retain the color and grayscale information of the images, which is beneficial for known class classification in certain domain shifts (\eg~from photo to art painting domain). However, using edge maps can greatly improve the model's capability of recognizing open-set classes, resulting in higher h-scores—a crucial metric in the context of OS-SDG.

\subsubsection{Overfiting Analysis}
\begin{figure}[H]
    \centering
    \includegraphics[width=0.9\linewidth]{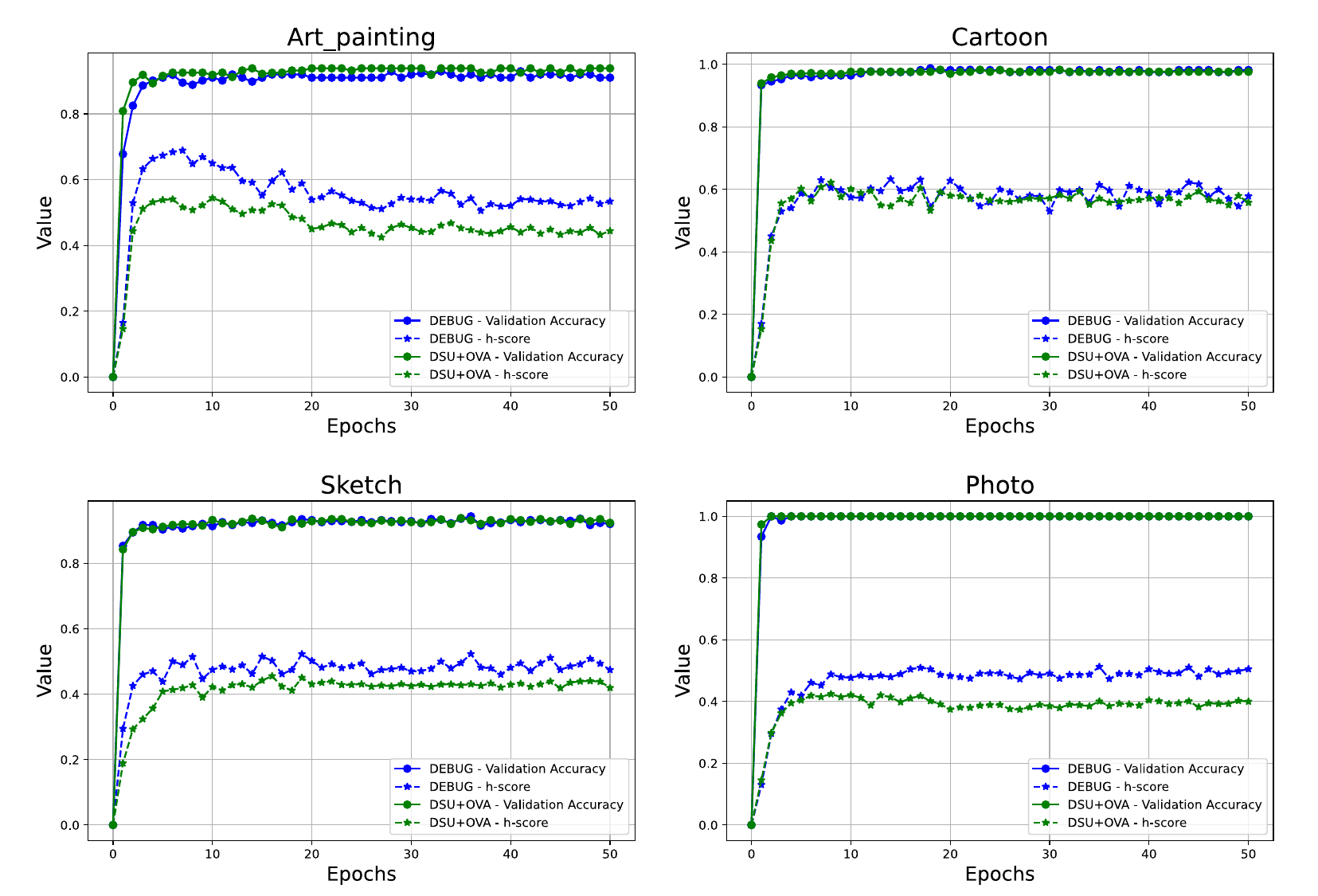}
    \vspace{-0.1in}
    \caption{Comparison of Validation Accuracy (on the source domain validation set) and h-score (on the remaining unseen domains) between DEBUG and the DSU+OVA baseline.}
    \label{fig:hs_per_domain}
\end{figure}

\begin{figure}[H]
    \centering
    \includegraphics[trim=0cm 0cm 0cm 1.5cm, clip,width=0.7\linewidth]{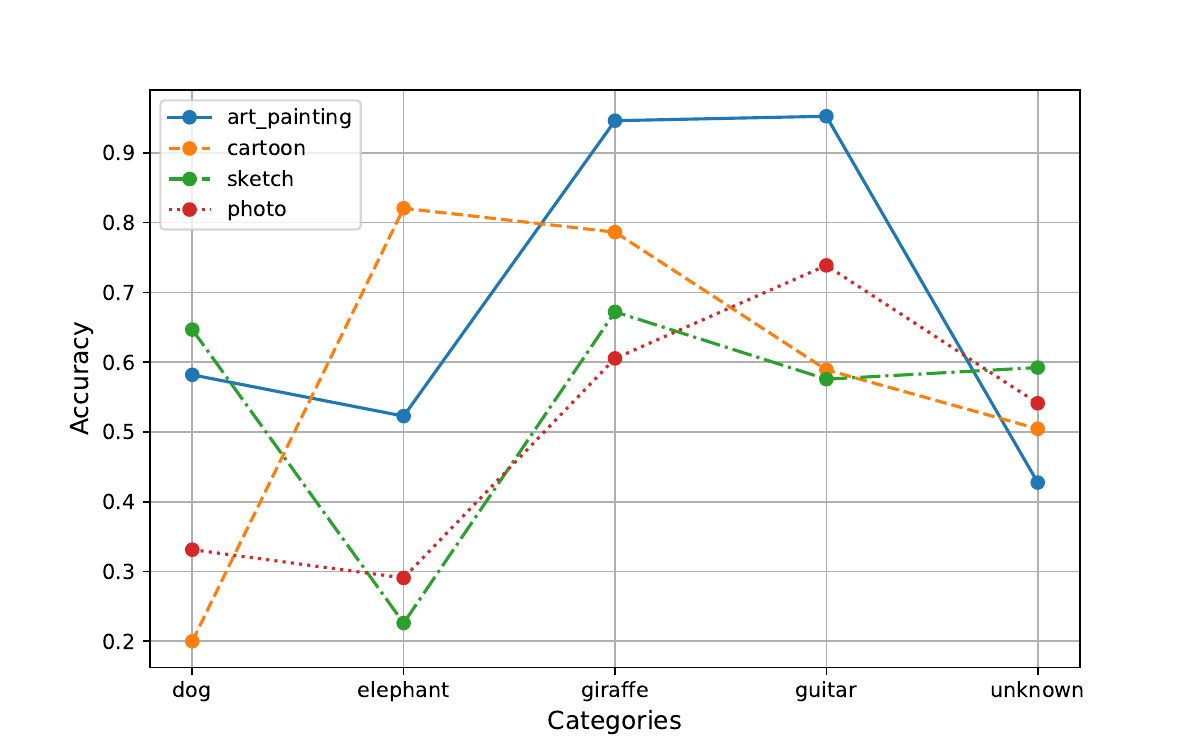}
    \vspace{-0.1in}
    \caption{Per-class accuracy on unseen domains using different source domains in the PACS dataset.}
    \label{fig:acc_per_class_per_domain}
\end{figure}

We utilize the PACS dataset to analyze overfitting in DEBUG due to its diverse domains. Figure \ref{fig:hs_per_domain} presents the h-score across various source domains for DEBUG and the DSU+OVA baseline, demonstrating that DEBUG consistently outperforms the baseline.
Figure \ref{fig:acc_per_class_per_domain} illustrates the per-class accuracy across different source domains, revealing certain imbalances. Specifically, the dog class in both the cartoon and photo domains, along with the elephant class in the photo and sketch domains, exhibit relatively lower accuracy, whereas other classes achieve higher accuracy.

\begin{figure*}[t] 
\centering
\includegraphics[width=0.7\textwidth]{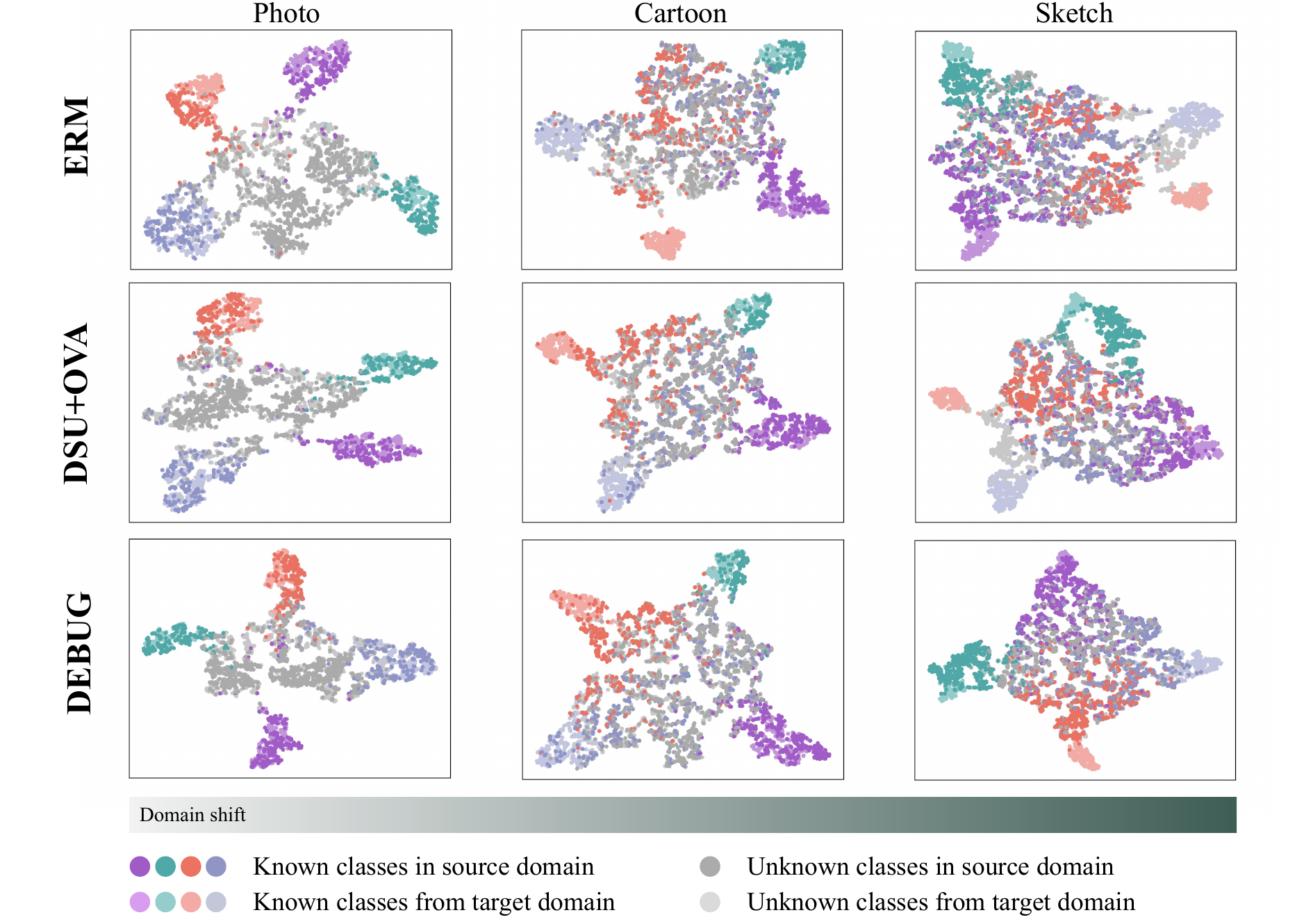} 
\vspace{-0.1in}
\caption{
The t-SNE visualizations of the encoded feature distribution in the source domain (Art Painting) and corresponding target domain for different methods on the PACS dataset.
}
\label{fig:tsne}
\end{figure*}

\subsubsection{Backbone Network Architecture}
We conduct experiments using two alternatives, the larger feature extractors: ResNet-50 and ResNet-101. 
The results are presented in Table \ref{tbl:model_arch}. As evident from the table, the performance of DEBUG significantly improves with the use of these feature encoders. These results demonstrate that DEBUG is compatible and effective when paired with different CNN feature encoders, further emphasizing the robustness and versatility of DEBUG.

\subsubsection{Visualization of Feature Distribution}
To further demonstrate the effectiveness of the proposed method, we employ t-SNE \cite{van2008tsne} to visualize the feature distribution of the known classes in the source domain, along with both the known and unknown classes within each target domain in the PACS dataset. The visualization results are presented in Figure \ref{fig:tsne}, where the source domain is art painting, and the target domains are photo, cartoon, and sketch, respectively.
The domain shift between the source and target domains increases gradually from left to right.
From the qualitative results of the ERM baseline, it is evident that with the increase in domain shift, the known class features in the target domain tend to move further away from their corresponding class centers in the source domain. Additionally, these features are more likely to intermingle with the features of other classes, indicating a higher level of confusion and difficulty in distinguishing between classes in the target domain.
The DSU+OVA method can partially mitigate feature drift (\ie~transition from art painting to cartoon) to some extent. However, its effectiveness diminishes significantly when faced with a substantial domain shift (\ie~from art painting to sketch).
In contrast, our proposed DEBUG method not only effectively aligns the features in the target domain when the domain shift is small, but also robustly prevents feature drift even in the face of a significant domain shift (\ie~from art painting to sketch). 
Moreover, we can observe that our DEBUG method yields more compact intra-class distributions and larger inter-class distribution boundaries for the known classes within the source domain when compared to the other two methods. 

\vspace{3pt}
\section{Conclusion}
In this paper, we proposed a novel domain expansion and boundary growth based approach, DEBUG, to address the OS-SDG task, where recognizing both the known and unknown classes in the unseen out-of-distributed domains poses significant challenges. In contrast to the only existing work that generates unstable or unreliable samples for the known and unknown classes, we discovered that compacting the intra-class distribution while expanding the inter-class distribution boundaries for the known classes within the source domain can effectively establish improved boundaries, not only between the known classes but also between the known and unknown classes in the unseen target domains; this, in turn, enhances the precision of recognizing out-of-distribution known classes, while also improving the ability to reject the unknown classes. 
Specifically, we designed the domain expansion through background suppression and global probabilistic-based style augmentation, while accomplishing intra-class distribution compaction by ensuring feature consistency across either the content or style perturbations. Furthermore, we achieved boundary growth by incorporating edge maps into multi-binary classifier training to enlarge the boundaries of the known classes, which facilitates the enhancement of the model capacity in open-set recognition. Extensive experiments on a variety of cross-domain datasets demonstrate the effectiveness of our method in the OS-SDG task.


\bibliographystyle{IEEEtran}
\small\bibliography{egbib}

\vspace{-0.2in}
\begin{IEEEbiography}
[{\includegraphics[width=1in,height=1.25in,clip,keepaspectratio]{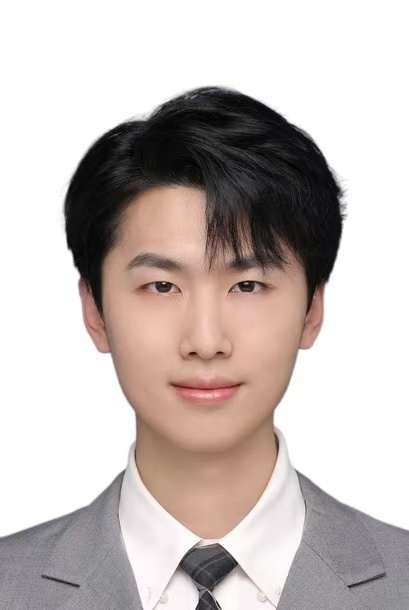}}]
{Pengkun Jiao} received the B.E. degree from East China Normal University, Shanghai, China, in 2021. He is currently pursuing
his Ph.D. degree in Computer Science at Fudan University. His research interests include domain adaptation, multi-media analysis and 3D vision.
\end{IEEEbiography}
\vspace{-0.2in}

\begin{IEEEbiography}
[{\includegraphics[width=1in,height=1.25in,clip,keepaspectratio]{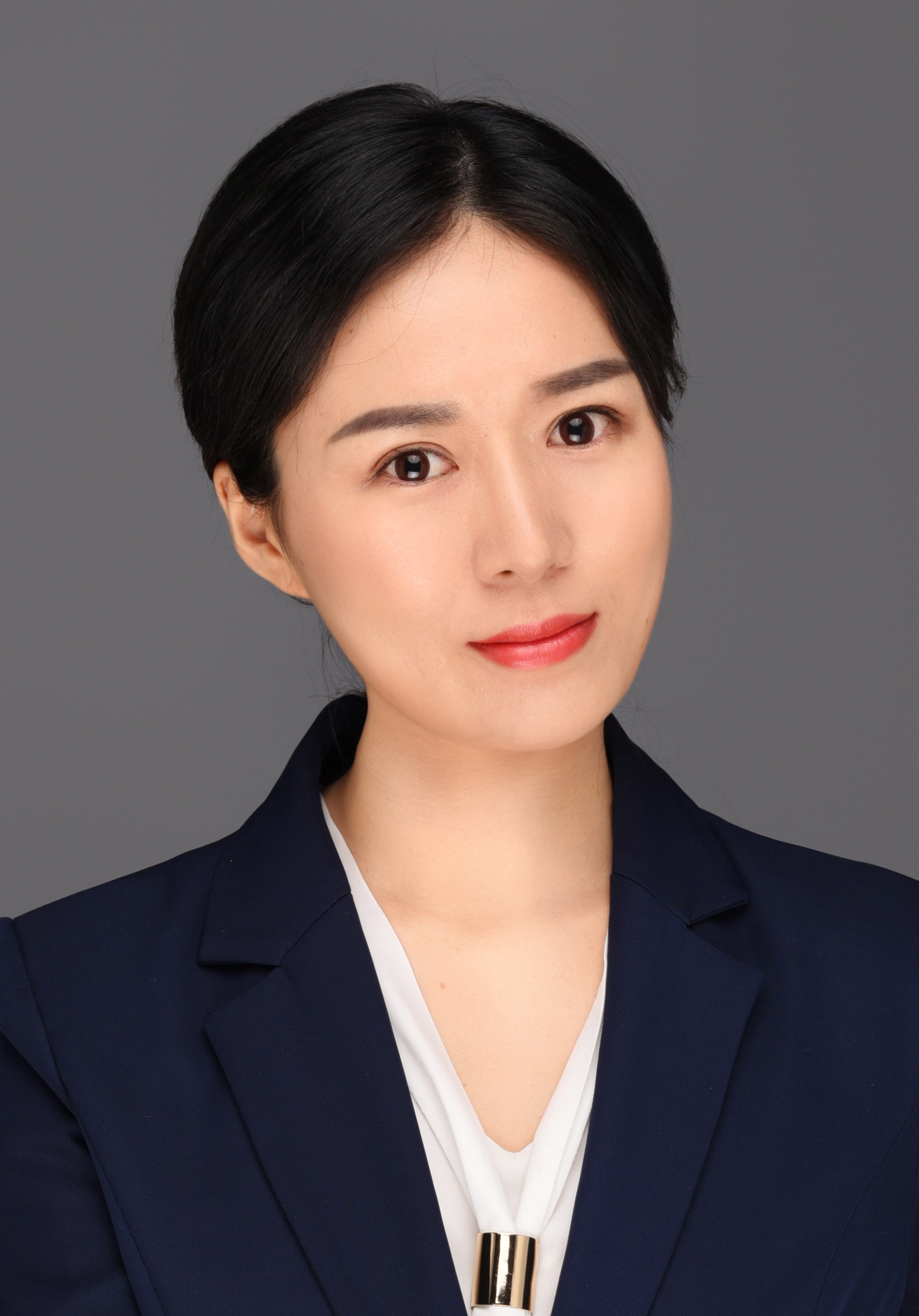}}]
{Zhao Na} is an assistant professor at Information Systems Technology and Design (ISTD) pillar, Singapore University of Technology and Design (SUTD). She received her PhD from National University of Singapore in 2021. Her research interests lie at the intersection of computer vision and machine learning. Her work has been published in top-tier conferences and journals in related fields, including CVPR, ICCV, ECCV, AAAI, Multimedia, ICRA, IROS, IJCV, TMM, and etc.
\end{IEEEbiography}
\vspace{-0.2in}

\begin{IEEEbiography}[{\includegraphics[width=1in,height=1.25in,clip,keepaspectratio]{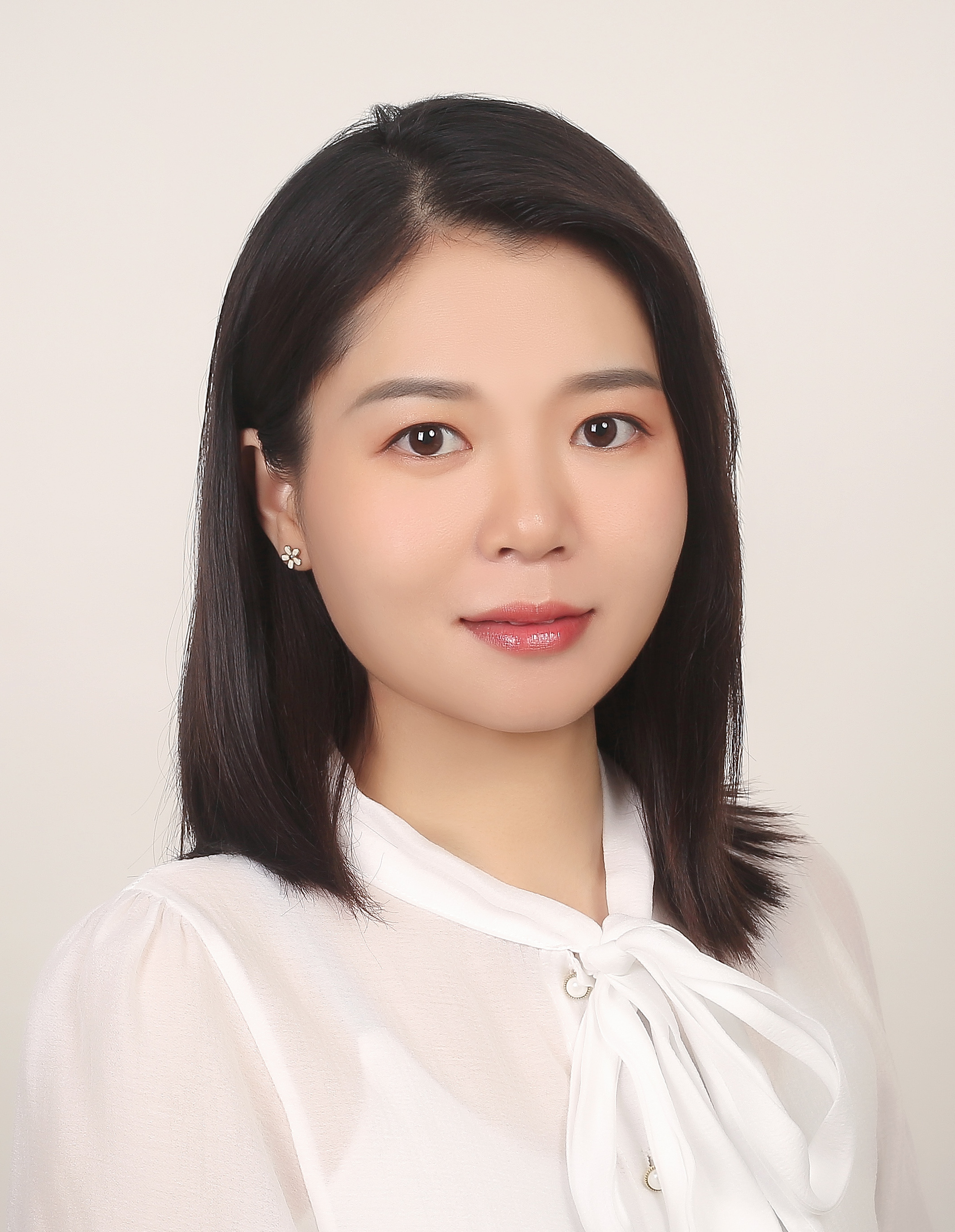}}]{Jingjing Chen} is now an Associate Professor at the School of Computer Science, Fudan University. Before joining Fudan, she was a postdoc research fellow at the School of Computing at the National University of Singapore. She received her Ph.D. degree in Computer Science from the City University of Hong Kong in 2018. Her research interest lies in diet tracking and nutrition estimation based on multi-modal processing of food images, including food recognition, and cross-modal recipe retrieval.
\end{IEEEbiography}
\vspace{-0.2in}

\begin{IEEEbiography}[{\includegraphics[width=1in,height=1.25in,clip,keepaspectratio]{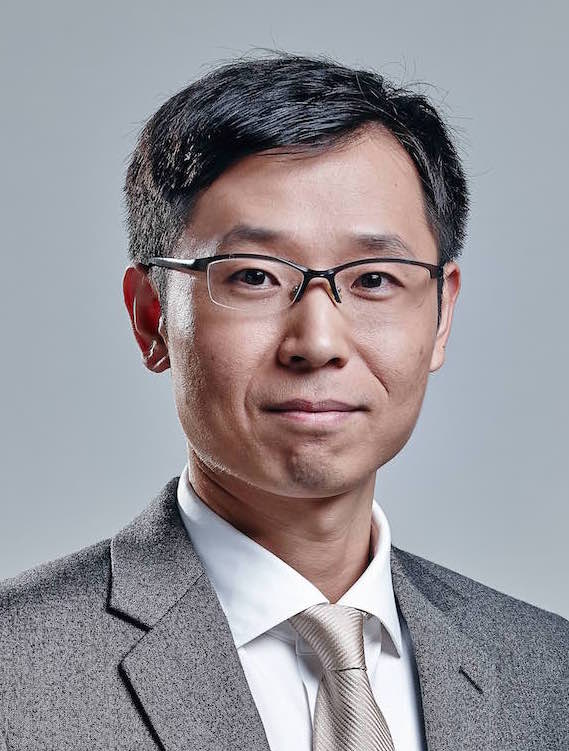}}]{Yu-Gang Jiang} received the Ph.D. degree in Computer Science from City University of Hong Kong in 2009 and worked as a Postdoctoral Research Scientist at Columbia University, New York, during 2009-2011. He is currently a Professor of Computer
Science at Fudan University, Shanghai, China. He has been elected as an IEEE Fellow in 2023. His research lies in the areas of multimedia, computer vision, and robust and trustworthy AI. His work has led to many awards, including the inaugural ACM China Rising Star Award, the 2015 ACM SIGMM Rising Star Award, the Research Award for Excellent Young Scholars from NSF China, and the Chang Jiang Distinguished Professorship appointed by the Ministry of Education of China.
\end{IEEEbiography}

\ifCLASSOPTIONcaptionsoff
  \newpage
\fi

\end{document}